\documentclass{IEEEtaes}
\usepackage{amsmath}
\usepackage{amsfonts}
\usepackage{amssymb}
\usepackage{dsfont}
\usepackage{xspace}
\usepackage{color,array,amsthm}

\usepackage[bottom]{footmisc}
\usepackage{graphicx}
\usepackage{subcaption}
\jvol{XX}
\jnum{XX}
\jmonth{XXXXX}
\paper{1234567}
\pubyear{2023}
\doiinfo{TAES.2023.Doi Number}

\begin{document}
\title{Contrastive Learning and Cycle Consistency-based Transductive Transfer Learning for Target Annotation }

\author{SHOAIB MERAJ SAMI}
\affil{ {LCSEE} Dept., West Virginia University, Morgantown, WV, USA} 
\author{MD MAHEDI HASAN}
\affil{ {LCSEE} Dept., West Virginia University, Morgantown, WV, USA} 
\author{NASSER M. NASRABADI}
\member{Fellow, IEEE}
\affil{ {LCSEE} Dept., West Virginia University, Morgantown, WV, USA}  

\author{RAGHUVEER RAO}
\member{Fellow, IEEE}
\affil{Army Research Laboratory, Adelphi, MD, USA}

\markboth{SAMI ET AL.}{TRANSDUCTIVE TRANSFER LEARNING FOR TARGET ANNOTATION}
\maketitle

\begin{abstract} Annotating automatic target recognition (ATR) is a highly challenging task, primarily due to the unavailability of labeled data in the target domain. Hence, it is essential to construct an optimal target domain classifier by utilizing the labeled information of the source domain images. The transductive transfer learning (TTL) method that incorporates a CycleGAN-based unpaired domain translation network has been previously proposed in the literature for effective ATR annotation. Although this method demonstrates great potential for ATR, it severely suffers from lower annotation performance, higher Fr\'echet Inception Distance (FID) score, and the presence of visual artifacts in the synthetic images. To address these issues, we propose a hybrid contrastive learning base unpaired domain translation (H-CUT) network that achieves a significantly lower FID score. It incorporates both attention and entropy to emphasize the domain-specific region, a noisy feature mixup module to generate high variational synthetic negative patches, and a modulated noise contrastive estimation (MoNCE) loss to reweight all negative patches using optimal transport for better performance. Our proposed contrastive learning and cycle-consistency-based TTL (C3TTL) framework consists of two H-CUT networks and two classifiers. It simultaneously optimizes cycle-consistency, MoNCE, and identity losses. In C3TTL, two H-CUT networks have been employed through a bijection mapping to feed the reconstructed source domain images into a pretrained classifier to guide the optimal target domain classifier. Extensive experimental analysis conducted on three ATR datasets demonstrates that the proposed C3TTL method is effective in annotating civilian and military vehicles, as well as ship targets. 
\end{abstract}


\section{INTRODUCTION}
Automatic target recognition (ATR) plays a vital role in numerous civilian and military applications, including target detection and vehicle classification~\cite{nasser_deeptarget}. It can contribute to reducing civilian casualties, aiding in rescue operations, and guiding land and maritime vehicles. Typically, ATR imagery is captured in various domains such as visible (VIS), mid-wave infrared (MWIR), long-wave infrared (LWIR), and synthetic aperture radar (SAR)~\cite{atr1,atr2_review,atr3,atr4}. Implementing ATR algorithms effectively requires a large amount of labeled data ~\cite{large_training}. However, in many ATR applications, we have labeled images in one domain (source) but no annotated images in the other domain (target). Manually annotating a large amount of target images is not only expensive but also time-consuming and cumbersome. Many semi-supervised and self-supervised algorithms~\cite{simclr,simclr2} have been proposed in recent literature, aiming to alleviate the manual annotation process for images. Moreover, transductive CycleGAN~\cite{TransductiveCycleGAN} was proposed to annotate unlabeled target domain (visible) images from the information of labeled source domain (MWIR) ATR images. However, this approach still encounters challenges such as low annotation performance and visual artifacts in the synthetic image due to the limitations of the unpaired image-to-image (I2I) translation network, specifically CycleGAN~\cite{cyclegan}, which fails to accurately transfer the domain-specific information. Therefore, to improve the transductive transfer learning (TTL) network performance, we replace the generator of CycleGAN-TTL by the proposed hybrid contrastive learning based unpaired domain translation (H-CUT) network. In addition, we retain the cycle-consistency in the TTL network to preserve the geometric structure and class information of the reconstructed images which also fed into a pretrained source domain classifier to backpropagate the cross-entropy loss. The proposed \textbf{c}ontrastive learning and \textbf{c}ycle-\textbf{c}onsistency-based \textbf{TTL} (\textbf{C3TTL}) network has two H-CUT networks ensuring two-way translation which captures the two distinct domain information accurately similar to DCLGAN~\cite{cut_dual_contrastive}. However, unlike DCLGAN, which employed similarity loss, we, in this paper utilize cycle-consistency loss.\\

The CUT~\cite{cut} network provides state-of-the-art performance in unpaired I2I translation scenarios. However, it has some issues, such as: i) randomly selecting queries, and ii) applying a similar contrastive force to all negative samples (patches). These issues are solved independently in~\cite{cut_qs_attn,cut_monce}. In this paper, we simultaneously employed these techniques with the myriad of synthetic negative patches generation, called hybrid-CUT (H-CUT) network. To summarize, the H-CUT network has three components i) query-selected attention (QS-Attention), ii) high variational negative patch generation, and iii) modulated contrastive noise contrastive estimation (MoNCE). The QS-Attention module avoids unnecessary modification during domain translation, emphasizing and maintaining higher-order descriptions about the shape and texture of the source domain images. The negative patches and their hardness play a vital role in contrastive learning within the CUT network~\cite{cut_monce}. In the contrastive learning literature, hard negative patch generation by mixup has proven effective techniques for better performance and faster convergence~\cite{hard_negative_Mochi}. Therefore, we have utilized both the original and synthetic negative patches in the H-CUT network. We use generated synthetic patches through a noisy feature mixup~\cite{noisy_feature_mixup} module. Furthermore, all the negative patches do not have a similar hardness compared to the query. Therefore, the multifaceted hardness of the negative patches needs to be considered for contrastive learning. We computed the hardness of negative patches using the optimal transport plan as reported in~\cite{cut_monce}. This modulated weighting-based NCE loss is called MoNCE loss. In the H-CUT network, we employ a hard weighting strategy in MoNCE calculation because it has benefits in unpaired I2I translation.
In our proposed C3TTL framework, we combine two H-CUT networks, one pretrained source domain classifier and another gradually learning target domain classifier. We optimize MoNCE, adversarial, and identity losses in the H-CUT network. After utilizing the H-CUT module in the TTL framework, the annotation performance improves significantly (7.06\%) in the DSIAC~\cite{dsiac} ATR dataset compared to CycleGAN-based TTL. The contribution of our work is summarized as follows:

\begin{enumerate}
 \item  We propose a new H-CUT network for the unpaired image-to-image translation task, an essential component of the proposed transductive transfer learning algorithm. The H-CUT network is a better unpaired I2I system by selecting query, re-weighting negative patches, and generating synthetic negative patches in the CUT network.
\item We also design a contrastive learning and cycle-consistency-based TTL (C3TTL) framework to annotate unlabeled target domain images. 
\item We exhaustively ablate our C3TTL framework by systematically removing cycle-consistency, NFM, and QS-Attention modules and replacing MoNCE loss by PatchNCE loss. The experimental result depicts that these components and losses are essential for better annotation performance and higher-quality synthetic images.
\item  Experiment on two military vehicle datasets and one ship target dataset demonstrates the effectiveness of the proposed C3TTL framework.
\end{enumerate}

Part of the material and experimental result was presented in the CycleGAN-TTL paper~\cite{TransductiveCycleGAN}. This current article extends in several ways: i) we propose the H-CUT as a one-way unpaired I2I domain translation network, ii) we present and explain each H-CUT and C3TTL network component. The architectural and implementation details of the proposed C3TTL method are depicted in this article. For reproducibility of the experiment, the source code will be published online; iii) we extend the experimental analysis presented in~\cite{TransductiveCycleGAN} multifaceted directions. First, we conduct experiments on three ATR datasets, evaluating the performance of C3TTL and CycleGAN-TTL. These datasets provide a wider annotation scope compared to others. Secondly, we perform ablation studies on each component of the proposed C3TTL method. Thirdly, we visualize synthetic images and Fr\'echet Inception Distance comparison among different variants of TTL. Fourthly, we compare the annotation performance of proposed TTL methods with several state-of-the-art baselines, such as SimCLR~\cite{simclr}, BYOL~\cite{byol}, SwAV~\cite{swav}, B-Twins~\cite{btwin}. We observe that C3TTL performance is better than those methods in semi-supervised settings.

The remainder of our work is developed as follows: Section~\ref{sec:related_work} presents a literature review related to our study; Section~\ref{sec:method} explains the proposed TTL network architecture and describes each of its modules; Section~\ref{sec:experiments} illustrates the effectiveness of the proposed network and each of its components with organized experiments; and finally, in Section~\ref{sec:conclusion}, we conclude our paper.

\section{Related Work}\label{sec:related_work}
This section will discuss the relevant work regarding automatic target recognition, transductive learning, and unpaired image translation network.

\subsection{Automatic Target Recognition}
The literature on automatic target recognition can be divided into two categories: (a) target detection and (b) target classification task. Different deep learning-based methods, such as Faster R-CNN~\cite{rcnn}, SSD~\cite{ssd}, YOLO~\cite{yolo}, etc., have been employed to detect civilian and military vehicles in different ATR datasets~\cite{nasser_deeptarget}. In~\cite{meta_domenic}, authors proposed Meta-UDA to detect unlabeled MWIR target images by using the labeled visible domain images in the DSIAC dataset~\cite{dsiac}. The ATR target classification algorithms can be divided into i) feature-based and ii) model-based approaches. The performance of the feature-based algorithm is lower compared to the model-based approaches. Convolutional neural networks (CNNs)~\cite{nasser_deeptarget}, generative adversarial networks (GANs)~\cite{semi-supervised_gan}, recurrent neural networks (RNNs)~\cite{rnn_lstm}, and autoencoders~\cite{autoencoder_deng} have been widely used for model-based target classification. The author~\cite{nasser_deeptarget} used a deep neural network to classify FLIR images in the Comanche (Boeing–Sikorsky, USA) dataset~\cite{comanche}. Ding et al.~\cite{noise_sar_ding} used different augmentation techniques, such as translation, speckle noise, and generating synthetic images with different poses to classify SAR images. Furthermore, Wang et al.~\cite{decouple_neural_noise} used a  dual-stage coupled network aggregating a despeckling subnetwork and a CNN-based classifier for classifying noisy SAR target images.

Few-shot learning is utilized in ATR to recognize target classes based on a limited number of labeled training samples. For instance, A few-shot learning method based on Conv-BiLSTM Prototypical Network ~\cite{few_shot_bilstm}, hybrid inference~\cite{hybrid_inference_wang2021}, and meta-learning~\cite{few_shot_meta} have been employed to classify the SAR images on MSTAR SAR dataset. Additionally, autoencoder and supervised constraint~\cite{autoencoder_deng} were exploited to classify the small samples in the ATR dataset. Also, in~\cite{semi-supervised_gan}, a semi-supervised method was proposed to classify the SAR images where the original and GAN-generated synthetic images were used to construct an ATR classifier.

\subsection{Transductive Transfer Learning}
Transductive transfer learning (TTL) is an effective tool for classifying unlabeled data in various target domains. For instance, Marcacini et al.~\cite{Marcacini2018-co} suggested a cross-domain aspect label propagation-based TTL method for opinion mining and sentiment analysis. TTL-based RNN was investigated to achieve domain adaptation from text to image for optical character recognition application~\cite{He2018-ql}. In~\cite{Zong2016-nb}, a sparse transductive transfer linear discriminant analysis model was proposed for speech emotion recognition in the wild. In~\cite{Yan2019-gi}, Yan et al. proposed a deep TTL framework for end-to-end cross-domain expression recognition. Their framework comprised two layers: a VGGFace16-Net-based feature extraction layer for extracting facial features from multi-view images, and a TTL module to eliminate the variation in feature distribution between the source and target domain images, and predict the class labels of the target images. The framework optimized cross-entropy loss and regression loss for the source and target domains. 

Furthermore, an inductive and supervised TTL approach~\cite{Kobylarz2020-qt} was employed for electromyographic gesture classification. In~\cite{Fu2017-bn,Moreo2021-bj} a TTL model was used for document and text classification using genetic programming, and distributional correspondence indexing. The authors~\cite{Luo2022-ld} utilized a transductive transfer learning-based approach for crop classification. A TTL network based on the deep forest classifier was utilized for cross-domain transfer learning and for measuring accuracy across different datasets, including MNIST, USPS, Amazon, DSLR, Webcam, and Caltech-256 datasets~\cite{Utkin2018-gg}.

\subsection{Unpaired I2I Domain Translation} 
The literature on unpaired domain translation is vast. The breakthrough in this field came with the CycleGAN paper~\cite{cyclegan}, which introduced a bijection mapping while preserving cycle-consistency. The cycle-consistency has also been employed in various aspects of different tasks~\cite{cyclegan_munit,drit,cyclegan_unit}. On the other hand, instead of relying on cycle-consistency loss, several works explored one-way unpaired image translation~\cite{cycle_no_1way1, cycle_no_1way2,cycle_no_1way3}. However, these approaches typically involve forming specific distance functions or calculating full image statistics. In contrast to these methods, contrastive unpaired I2I translation (CUT)~\cite{cut} improves image quality and convergence by replacing cycle-consistency with information maximization between input and output patches from the source and target domain.

The CUT network and its variants have been widely employed for unpaired I2I translation. However, a crucial concern with the CUT network is that it uniformly contrasts all the negative patches without considering their similarity to the anchor. To address this issue, Zhan et al.~\cite{cut_monce} introduced the MoNCE loss, which incorporates optimal transport to collaboratively re-weight all negative samples across different objectives. Hard negative sample generation~\cite{cut_negcut} and learned self-similarity~\cite{cut_flsesim} based CUT methods have also been explored in this field. Attention for specifying the domain-specific and domain-invariant information has also been employed in the CUT network~\cite{cut_qs_attn}. DCLGAN~\cite{cut_dual_contrastive} used two-way translation by considering the CUT network. Furthermore, the denoising diffusion model has been successfully exploited in the unpaired image-to-image translation method~\cite{diffusion_egsde,diffusion_sddm}.

\section{Methodology}\label{sec:method}
The proposed C3TTL model comprises two H-CUT networks and two classifiers. Each H-CUT network has a separate generator and a discriminator. In this network, the source domain X contains a set of images, denoted as x, while the target domain Y contains another set of images, denoted as y. We express them $X = \{x \in X\}$ and $Y = \{y \in Y\}$. We intend to discern two mappings $G_{XY}:X\rightarrow Y$, which translates images from domain $X$ to domain $Y$, and $G_{YX}:Y\rightarrow X$, which performs the reverse translation. 

The C3TTL model has two generators $G_{XY}$, $G_{YX}$ and two discriminators $D_{X}$, $D_{Y}$. $D_{X}$ and $D_{Y}$ assure that the translated images stay in the proper domain. The first half of each generator is denoted as the encoder ($G_{Enc}$), and the second half is denoted as the decoder ($G_{Dec}$).

For each H-CUT~\cite{cut} network, a stack of features is extracted from different layers of the encoder and projected through a two-layer MLP head (H). Let the first H-CUT network transforms source domain images to target domain images. We utilize $G_{XY(Enc)}$ for encoder and H$_{1}$ for embedding. On the other hand, we also employ the second H-CUT network to transform target domain images into source domain images. It utilizes the $G_{YX(Enc)}$ encoder and H$_{2}$ embedding. We also employ cycle-consistency for preserving the geometric structure of reconstructed source domain images which are fed to a pretrained classifier to guide the C3TTL model. This network has two classifiers ($C_{source}$ and $C_{target} $).\\
\begin{figure*} [ht]
\begin{center}
\begin{tabular}{c} 
\includegraphics[scale = .6]{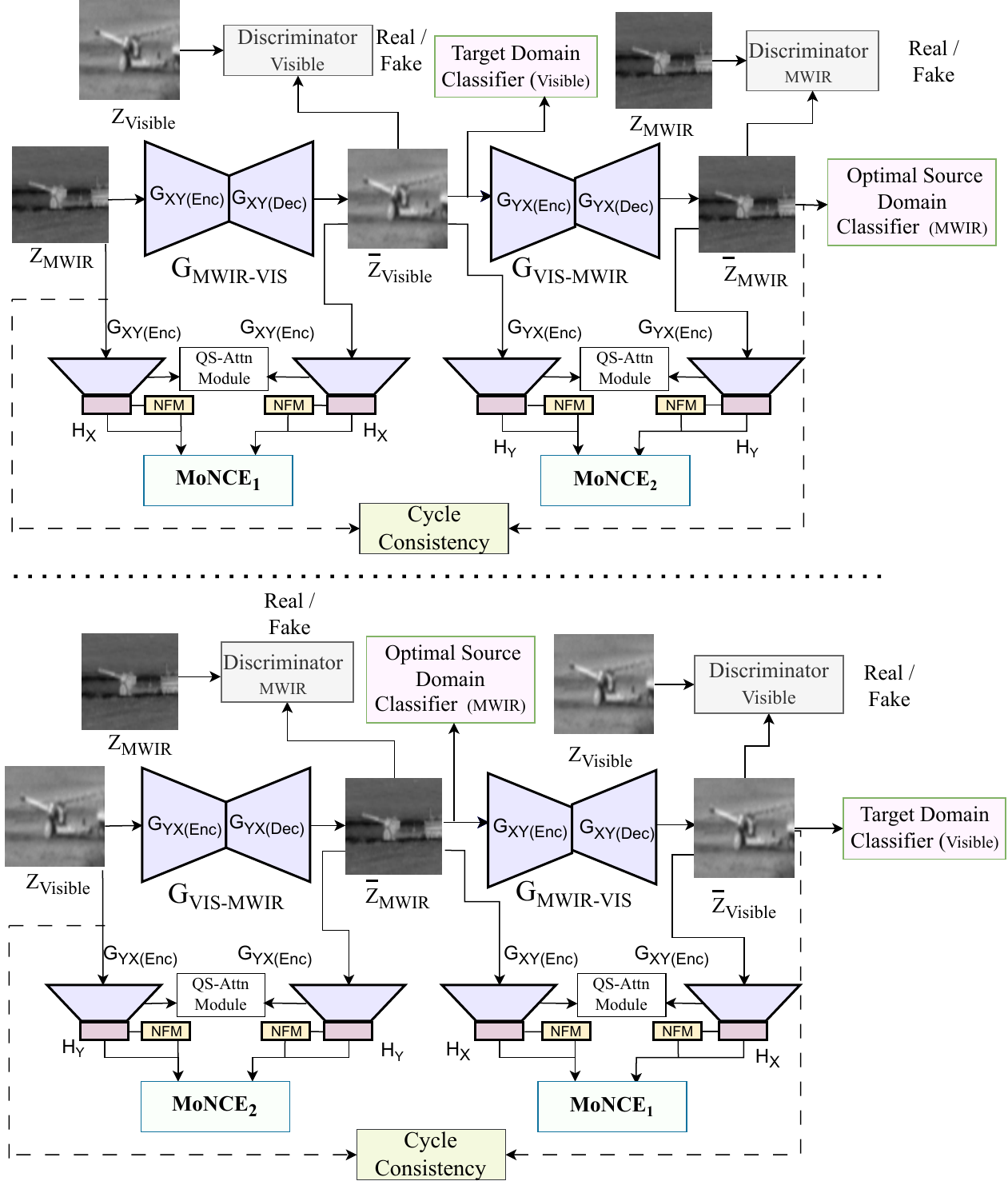}
\end{tabular}
\end{center}
\caption[example] 
{Overview of our proposed C3TTL framework. It consists of two generators: one generator ($G_{MWIR-Vis}$) for transforming source domain images (MWIR) to the target domain (Visible) and the second generator ($G_{Vis-MWIR}$) for transforming synthetic target domain images (Visible) back into the source domain (MWIR). Each generator has its own discriminator ($D_{Vis}$ and $D_{MWIR}$) and a dedicated CNN-based ATR classifier ($C_{Vis}$ and $C_{MWIR}$). Furthermore, the hybrid contrastive unpaired I2I translation (H-CUT) network is the building block of the C3TTL. The H-CUT network comprises QS-Attention, noisy feature mixup (NFM) modules, and MoNCE loss.  \label{fig:C3TTL}}
\end{figure*} 
An overview of our proposed C3TTL framework is depicted in~\figurename ~\ref{fig:C3TTL}. Our proposed network combines five losses: adversarial loss, MoNCE loss, identity loss, cycle-consistency loss, and categorical cross-entropy loss which are described in the following subsections. 
The details of our C3TTL and CycleGAN-TTL frameworks are also described below.
\subsection{Adversarial Loss}
A generative adversarial network (GAN)~\cite{gan} is a powerful tool for different types of image generation and image translation. The building blocks of a GAN architecture are a generator and a discriminator. The GAN generates images from a random variable, $z$, where the generator function can be represented as $G: z\to y$. On the other hand, the discriminator function, $D$, is designed to distinguish between the real image distribution and generated image distribution. In GAN, the generator and the discriminator compete with each other as a two-player mini-max game~\cite{gan}. 
\\
The first and second H-CUT networks optimize $\mathcal{L}(G_{XY}, D_Y)$, and $\mathcal{L}(G_{YX}, D_X)$, respectively. The summation of the above-mentioned losses is the adversarial loss ($\mathcal{L}_{GAN}$) of the C3TTL network:

\begin{equation*}
\begin{split}
\mathcal{L}(G_{XY}, D_Y)\!&=\! \mathbb{E}_{x \sim p_{data}(x)}\log (1 \!-\! D_Y(G_{XY}(x))) \\
&+ \mathbb{E}_{y \sim p_{data}(y)}\log D_Y(y), 
\end{split}
\end{equation*}
\begin{equation*}
\begin{split}
\mathcal{L}(G_{YX}, D_X)\!&=\!  \mathbb{E}_{y \sim p_{data}(y)}\log (1 \!-\! D_X(G_{YX}(y))) \\
&+ \mathbb{E}_{x \sim p_{data}(x)}\log D_X(x), 
\end{split}
\end{equation*}

\begin{equation} \label{eq:2}
\mathcal{L}_{GAN} = \min_{G} \max_{D} \mathcal{L}(G_{XY}, D_Y) + \mathcal{L}(G_{YX}, D_X).
\end{equation}
\subsection{Cycle-Consistency Loss}

The CycleGAN~\cite{cyclegan} introduced cycle-consistency loss to enforce bijection mappings in unpaired cross-domain I2I transfers, thereby preserving semantic information during translation. The cycle-consistency loss can be denoted as $\mathcal{L}_{Cycle}$:

\begin{equation}\label{eq:3}
\begin{aligned}
\mathcal{L}_{Cycle} = & \mathbb{E}_{x\sim p_{\text{data}}(x)}\left( \left\|G_{YX}(G_{XY}(x))-x  \right\|_{1} \right)\\ +
&\mathbb{E}_{y\sim p_{\text{data}}(y)}\left( \left\|G_{XY}(G_{YX}(y))-y  \right\|_{1} \right).
\end{aligned}
\end{equation}

\subsection{Identity Loss}
To prevent unnecessary changes to the generators, we use identity mapping regularizers~\cite{cyclegan}. This identity loss can be expressed as follows:
\begin{equation}\label{eq:4}
  \begin{aligned}
  \mathcal{L}_{Identity} = \mathbb{E}_{y\sim p_{\text{data}}(y)}\left( \left\|G_{XY}(y)-y  \right\|_{1} \right)\\+\mathbb{E}_{x\sim p_{\text{data}}(x)}\left( \left\|G_{YX}(x)-x  \right\|_{1} \right) .
    \end{aligned}
\end{equation}

\subsection{PatchNCE Objective}
The CUT~\cite{cut} introduced a method to maximize the mutual information between the input patch and the corresponding output patch to preserve the semantic content in an unpaired I2I translation scheme by employing a contrastive learning framework. It consists of an adversarial loss and a PatchNCE loss. It actually maximizes the mutual information, $I(X,Y)=H(X)-H(X|Y)$, which is equivalent to minimizing the conditional entropy $H(X|Y)$, a similar goal to cycle-consistency. The PachNCE objective can be denoted as:
\begin{equation}\label{eq:6_patch}
\small 
\begin{aligned}
\mathcal{L}_{\substack{{Patch}\\{NCE}}}(X,\bar{Y}) = -\sum_{i=1}^N{log} [{\frac{{e^{(\bar{y}_{i}.x_{i}{/\tau)}}}}{e^{(\bar{y}_{i}.x_{i}{/\tau)}}+\sum{^{N}_{\substack{{j=1}\\_{j\not = {i}}}}}{e^{(\bar{y}_{i}.x_{j}{/\tau)}}}}}],
\end{aligned}
\end{equation}
where $\tau$ is a temperature parameter and $\bar{Y}$, $X$ are the generated target domain and ground truth images, respectively. $X =[x_1,x_2,....,x_N]$ and $\bar{Y} =[\bar{y}_1,\bar{y}_2,....,\bar{y}_N]$ are encoded feature vectors from 1st, 4th, 8th, 12th and 16th layers of the encoder. Afterward, the features passed through a two-layer MLP network~\cite{cut,cut_monce}. In the above equation, N-class classification is performed, where the anchor provides the same contrastive force on the $N-1$ negative samples that can be too strict and detrimental. To address this issue, modulated contrast NCE loss has been proposed~\cite{cut_monce}. 

\subsection{Modulated Contrast NCE Objective}
\begin{figure} [h!]
\begin{center}
\begin{tabular}{c}

\includegraphics[scale = 0.077]{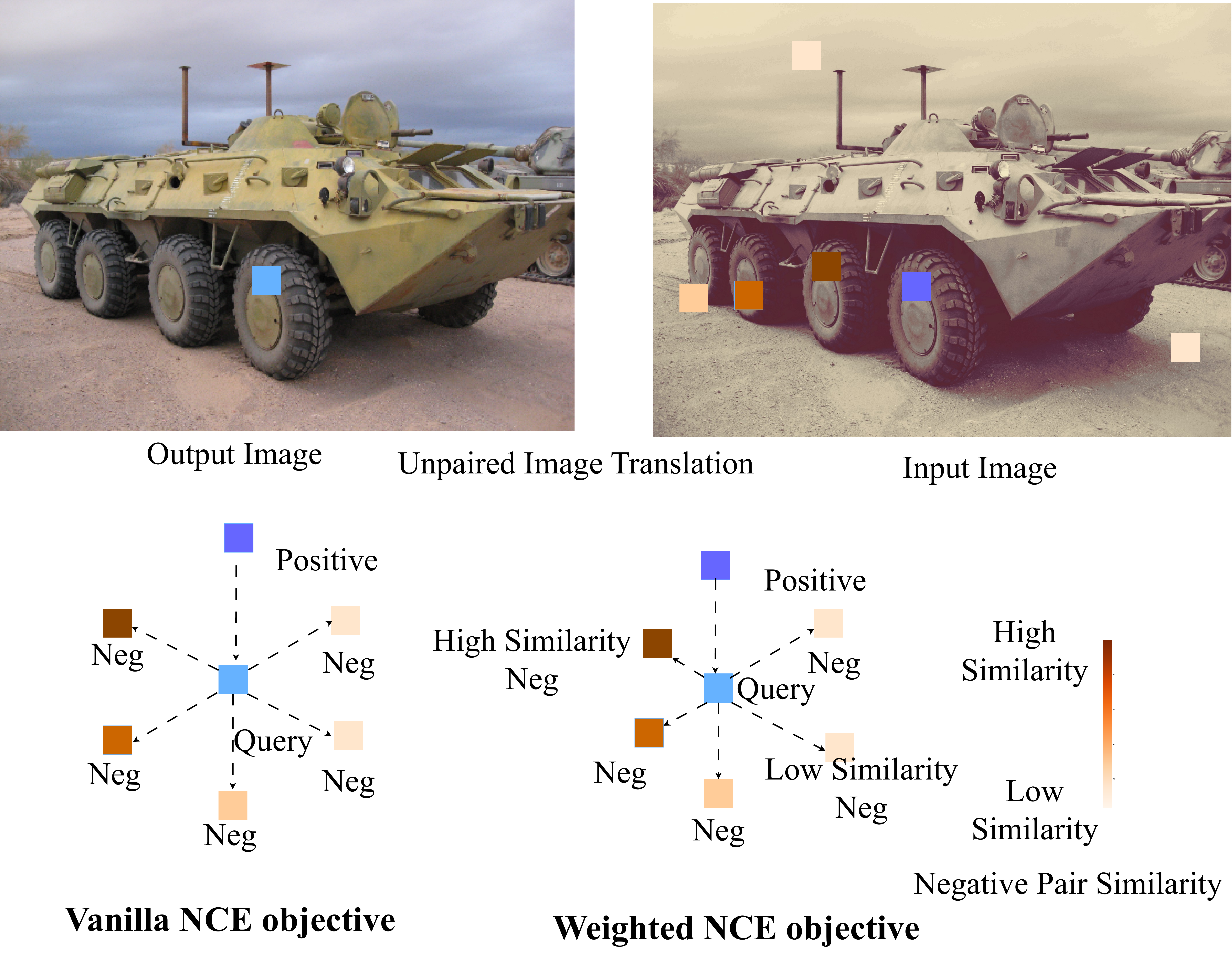}
\end{tabular}
\end{center}
\caption[example] 
{ \label{fig:monce1} 
Vanilla and weighted contrastive objective~\cite{cut_monce}.}
\end{figure}

\begin{figure} [h]
\begin{center}
\begin{tabular}{c} 
\includegraphics[scale = 0.075]{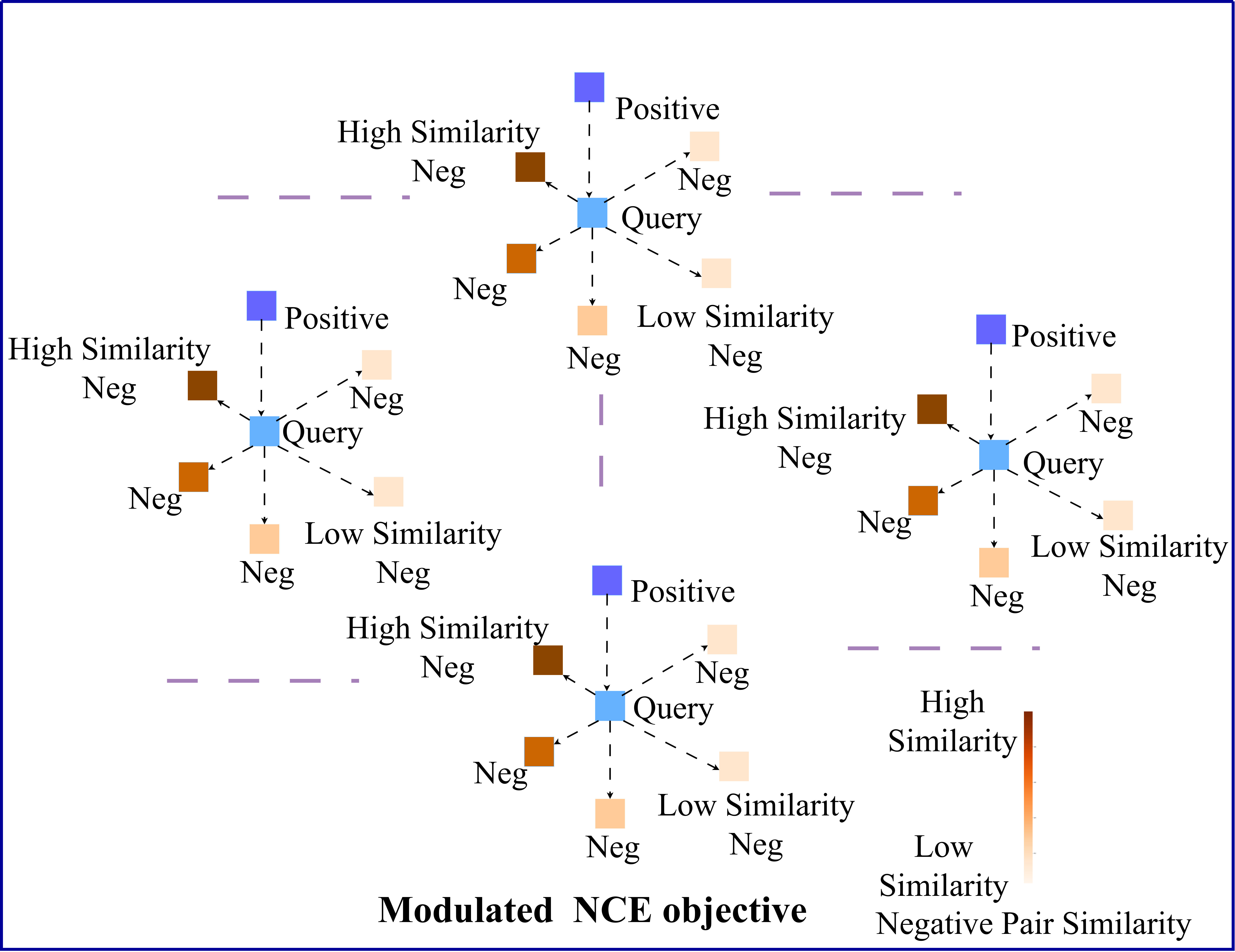}
\end{tabular}
\end{center}
\caption[example] 
{ \label{fig:monce2} 
Modulated contrastive objective~\cite{cut_monce}.}
\end{figure}

In the literature of contrastive learning, the hardness of negative samples has been adequately investigated~\cite{hard_negative_iclr,cut_negcut,hard_negative_Mochi}. In contrastive learning, the application of hard negatives has been proven to facilitate the learning of data representations~\cite{hard_negative_iclr}. Additionally, in the field of unpaired image translation, the hardness of negative patches is defined by considering their similarity to the query~\cite{cut_monce}. The similarity between a negative sample $x_j$ and an anchor $ \bar{y}_i$ is defined by the hard negative weighting, as shown in Eq.~\ref{eq:71}:

\begin{equation}\label{eq:71}
\begin{aligned}
w_{ij} = {\frac{{e^{(\bar{y}_{i}.x_{i}{/\beta)}}}}{\sum{^{N}_{\substack{{j=1}}}}{e^{(\bar{y}_{i}.x_{j}{/\beta)}}}}},
\end{aligned}
\end{equation}

\noindent where $\beta$ is the weighting temperature parameter.
The modulate contrast NCE loss objective employs reweighing strategies by enforcing the constraint defined by Eq.~\ref{eq:72}:
\begin{equation}\label{eq:72}
\begin{aligned}
{\sum^{N}_{i=1}}w_{ij}= 1, 
 {\sum^{N}_{j=1}}w_{ij}= 1  ;   i,j \in [1,N].
\end{aligned}
\end{equation}
Equation (\ref{eq:73}) is the optimal transport ~\cite{Optimal_transport} for this algorithm which also considers the Eq. (\ref{eq:72}) as a constraint:
\begin{equation}\label{eq:73}
\begin{aligned}
\min_{w_{ij}, i,j \in [1,N] } [\sum^{N}_{i=1}\sum^{N}_{\substack{{j=1}\\j\neq i}}w_{ij}.{e^{\bar{y}_{i}.x_{j}{/\tau}}}],
\end{aligned}
\end{equation}

\begin{equation}\label{eq:74}
\begin{aligned}
\min_{T}\langle C,T \rangle  \; \;    s.t  \; \;\langle T\overrightarrow{1}\rangle =1, \; \langle T^{T}\overrightarrow{1}\rangle =1,
\end{aligned}
\end{equation}
\noindent where $\langle C,T \rangle$ is the inner product of the cost matrix ($C$) and transport plan ($T$). In the unpaired I2I network, the cost matrix is ${e^{\bar{y}_{i}.x_{j}{/\beta}}}$ where $i \neq j$; if $i=j$ then $C_{ij}=\infty$. The Sinkhorn~\cite{Sinkhorn_original} algorithm is applied to Eq.~\ref{eq:74} for calculating the optimal transport plan. Furthermore, similar to the PatchNCE, the MoNCE method also incorporates multiple layers of the encoder features for contrastive learning. The examples of vanilla and modulated contrast are depicted in  \figurename ~\ref{fig:monce1}. and \figurename ~\ref{fig:monce2}. The MoNCE objective($\mathcal{L}_{MoNCE}$) can be expressed as:
\begin{equation}\label{eq:6_patch}
\small 
\begin{aligned}
\mathcal{L}_{MoNCE}=-\sum_{i=1}^N{log} [{\frac{{e^{(\bar{y}_{i}.x_{i}{/\tau)}}}}{e^{(\bar{y}_{i}.x_{i}{/\tau)}}+Q(N-1)\sum{^{N}_{\substack{{j=1}\\_{j\not = {i}}}}}{w_{ij}}{e^{\bar{y}_{i}.x_{j}{/\tau}}}}}],
\end{aligned}
\end{equation}
where $Q$ denotes the weight of negative terms in the denominator and typically $Q=1$.

\begin{figure*} [h!]
\centering 
\begin{center}
\begin{tabular}{c} 
\includegraphics[scale = 0.65]{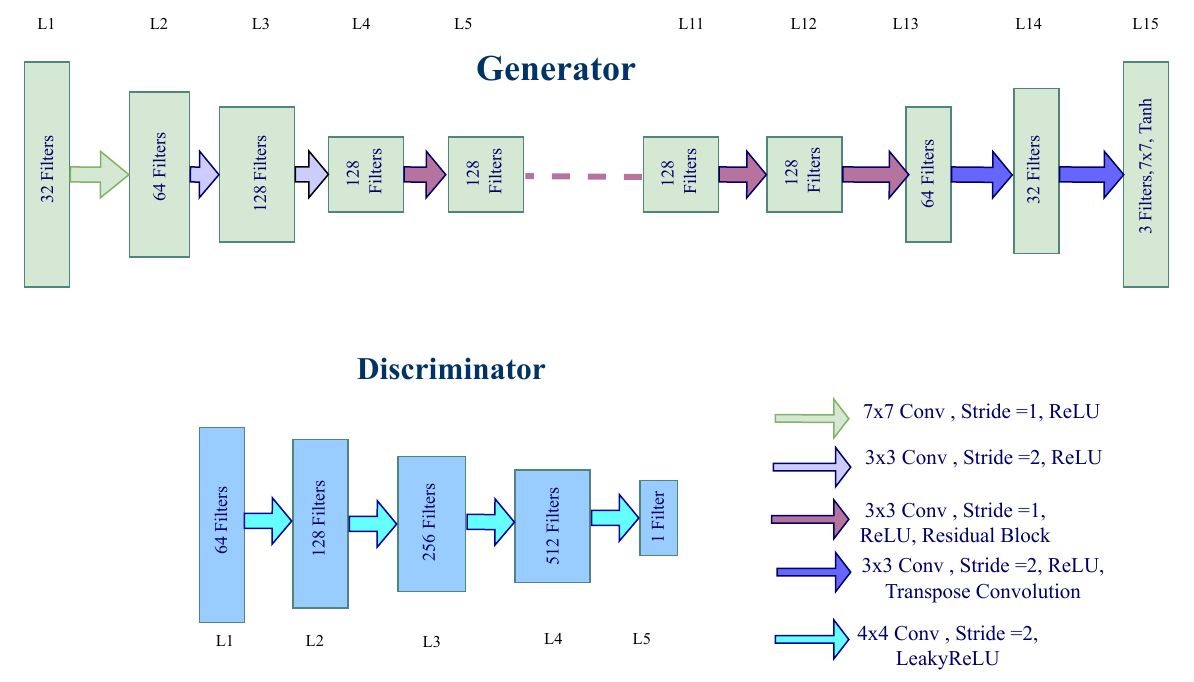}
\end{tabular}
\end{center}
\caption[example] 
{ \label{fig:example2} 
Block diagram of the generator and discriminator architecture in CycleGAN as well as CUT~\cite{TransductiveCycleGAN}.}
\end{figure*} 

\subsection{Query-Selected Attention-based CUT}
\begin{figure} [h]
\begin{center}
\begin{tabular}{c} 
\includegraphics[scale = 0.076]{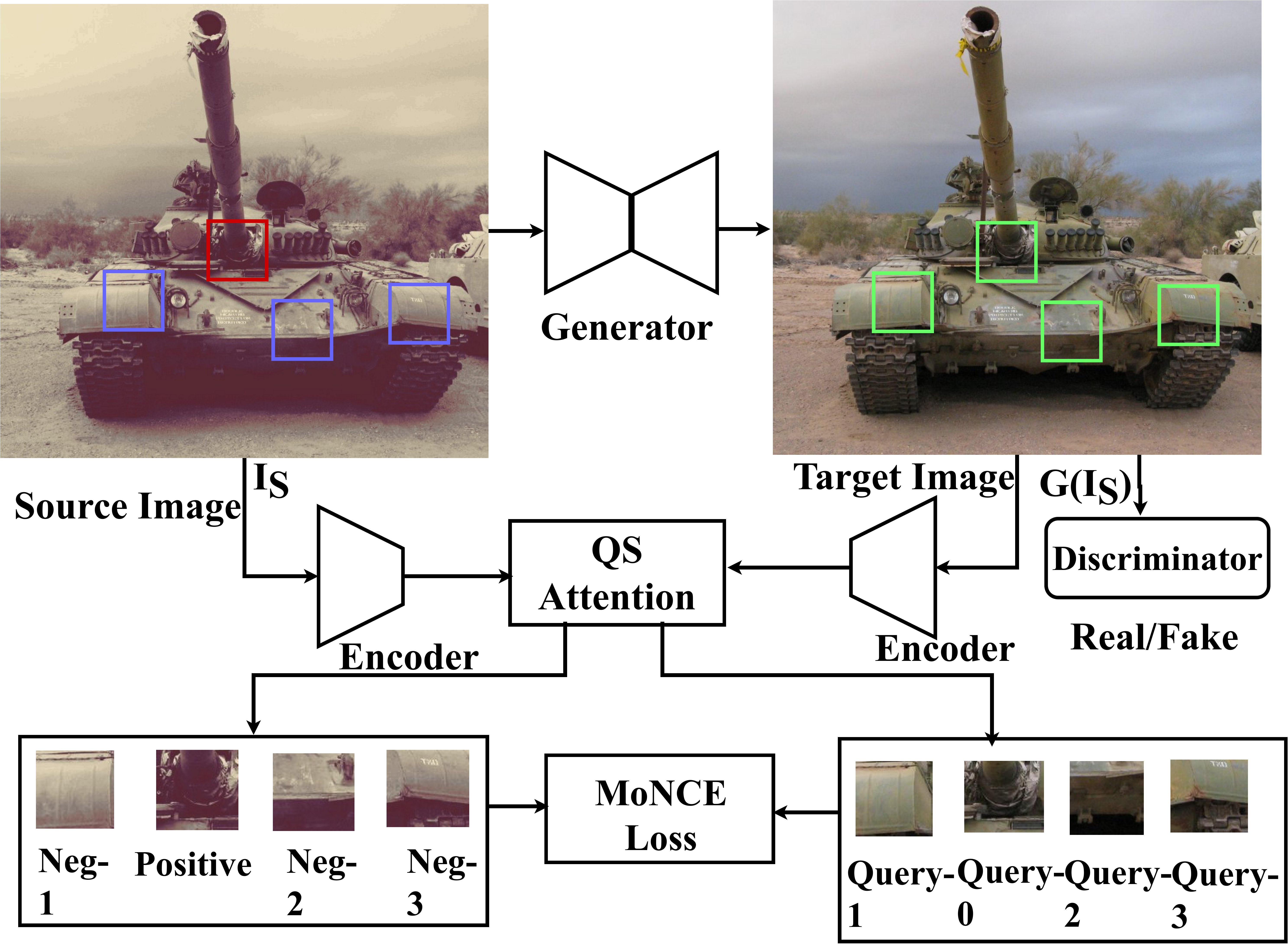}
\end{tabular}
\end{center}
\caption[example] 
{\label{fig:qs-attention} 
The query-selected attention-based H-CUT architecture~\cite{cut_qs_attn}.}
\end{figure}

The CUT network selects queries from random locations within the generated images; however, this approach may not always serve the intended purpose of domain translation. This is because not all locations of the image patches contain similar importance of domain-related information. The AttnGAN~\cite{attngan} guides the generator of the GAN to translate the relevant area of the images by using a learnable foregrounded mask. The F-LSeSim~\cite{cut_flsesim} method, a variant of CUT, uses self-similarity in the local regions of the image and employs the NCE loss. All the above-mentioned algorithms can not properly identify the domain-relevant patches. The query-selected attention (QS-Attn)~\cite{cut_qs_attn} overcomes this issue by using both attention and entropy for query selection and incorporates it into the CUT network.
For this purpose, the QS-Attention reshapes the source domain images $F_x$ $\in$ $\mathbb{R}^{H\times W \times C}$ into a two-dimensional matrix $Q$ $\in$ $\mathbb{R}^{HW\times C}$ and its transpose $K$ $\in$ $\mathbb{R}^{C\times HW}$. The $Q$ and $K$ are multiplied, and then a softmax normalization is applied row-wise, resulting in the generation of an attention matrix $A_{g}$ $\in$ $\mathbb{R}^{HW\times HW}$. The entropy is calculated to the attention matrix $A_{g}$. This scenario is illustrated in Eq.~\ref{eq:81}:
\begin{equation}\label{eq:81}
\begin{aligned}
 H_{g}(i) = - \sum^{HW}_{j=1}A_{g}(i,j)\log A_{g}(i,j),
\end{aligned}
\end{equation}
\noindent where $i$, $j$ denote the indices of query ($Q$) and key ($K$) metrics. For selecting the relevant query, the entropy matrix $H_g$ is sorted by ascending order and the smallest $N$ row is chosen. This calculation is performed only on the features of source domain images. The final size of the global QS-Attention matrix is $A_{QS}\in$ $\mathbb{R}^{N\times HW}$. The reduced $A_{QS}$ is applied on both the real and synthetic image to capture the global relation and higher-order delineations, such as texture and shape. The QS-Attention-based H-CUT network is illustrated in the \figurename~\ref{fig:qs-attention}.
 
\subsection{Generation of Synthetic Patches in the H-CUT Network}
The utilization of harder negative samples in the contrastive loss function is crucial for faster and better learning~\cite{hard_negative_iclr}. Additionally, hard negative mixing by MoCHi~\cite{hard_negative_Mochi}, has proven to improve performance and lead to faster convergence in contrastive learning. Moreover, data mixing can be performed at feature or pixel levels~\cite{contrastive_mixup_feature_pixel1, contrastive_mixup_feature_pixel2}. To improve the generalization and robustness of deep neural networks, various techniques have been proposed, including manifold mixup~\cite{contrastive_mixup_feature_pixel1}, CutMix, Puzzle Mix, and noisy feature mixup~\cite{noisy_feature_mixup}. The noisy feature mixup (NFM) has outperformed the other feature mixing methods in terms of robustness and smoothness of the decision boundary of the neural network model~\cite{noisy_feature_mixup}.

\subsubsection{Noisy Feature Mixup}
NFM combines the features of DNN and injects additive and multiplicative noise to improve generalization. If noise injection is not used, NFM can be described as a form of the manifold mixup~\cite{contrastive_mixup_feature_pixel1} method. In this paper, we only consider the feature-level NMF. The equation for NFM is presented by Eq.~\ref{eq:91}:
\begin{equation}\label{eq:91}
\begin{aligned}
\tilde\tilde{g_k} = (\mathds{1} +\sigma_{mult}\xi_{k}^{mult}) \odot {M}_{\lambda}(g_k(x),g_k(x')) + \sigma_{add}\xi_{k}^{add},
\end{aligned}
\end{equation}
\\where ${M}_{\lambda}(g_k(x),g_k(x')) = \lambda*g_k(x) +(1-\lambda)*g_k(x')$ and $\lambda \sim Beta(\alpha,\beta)$.
Here $\xi_{k}^{add}$ and $\xi_{k}^{mult}$ are $\mathbb{R}^{d_{k}}$ valued independent random variables that model the additive and multiplicative noise respectively, and $\sigma_{add}$, $\sigma_{mult}$ $ \geq 0$ are the specific noise levels. Furthermore, $g_k(x)$ and $g_k(x')$ are two negative patches, and after noisy feature mixup, we get $\tilde\tilde{g_k}$ as a synthetic negative patch. Additionally, $\odot$ denotes Hadamard product, and $\mathds{1}$ denotes the vector with all the components values equal one. 

\subsection{Source and Target Domain Classifiers}
The TTL network shown in \figurename~\ref{fig:C3TTL} has source and target domain ATR classifiers. In our experiment, the architecture of both classifiers is ResNet-18~\cite{resnet}. The skip connection in the ResNet-18 architecture makes it possible to build a deeper network and helps to backpropagate loss more efficiently. In the proposed method, we optimize the cross-entropy loss ($\mathcal{L}_{CE}$) of the classifiers:
\begin{equation}\label{eq:6_ce}
  \begin{aligned}
   \mathcal{L}_{CE} = -\sum_{c=1}^{C}y_{o,c}\log(p_{o,c}),
    \end{aligned}
\end{equation}
where $y_{o,c}$ is the true probability and $p_{o,c}$ is the predicted probability of the target chip in the training data, and the total number of classes is denoted by $C$.

\subsection{Transductive CycleGAN}
Our previously proposed transductive CycleGAN~\cite{TransductiveCycleGAN} consists of a CycleGAN and two classifiers. The architecture of the transductive CycleGAN generators and discriminators is illustrated in \figurename~\ref{fig:example2}. The generator consists of fractionally-strided convolutions, regular convolutions, and residual blocks.  

Furthermore, the source domain classifier is assumed to be well-trained (called the optimal source domain classifier), and its weights are fixed. During the training, the target domain classifier weights are initialized by the source domain classifier weights. The CycleGAN-TTL network is optimized by minimizing the adversarial loss, cycle-consistency loss, identity loss of the CycleGAN, and the categorical cross-entropy loss of the source and target domain classifiers. It is noteworthy to mention that although the source domain classifier loss is backpropagated to the transductive network, the weights of the source domain classifier are not updated during the training process of the transductive network.

\subsubsection{CycleGAN-TTL Objective}
The total CycleGAN loss ($\mathcal{L}_{CycleGAN}$) is the summation of the adversarial loss, cycle-consistency loss, and identity loss: 
\begin{equation}\label{eq:13}
  \begin{aligned}
  \mathcal{L}_{CycleGAN} =\lambda_{a}* \mathcal{L}_{GAN} +\lambda_{b}* \mathcal{L}_{Cycle} + \lambda_{c}*\mathcal{L}_{Identity},
    \end{aligned}
\end{equation}
here, $\lambda_{a}$, $\lambda_{b}$, and $\lambda_{c}$ are the hyper-parameters.
The total loss of the transductive CycleGAN model is $\mathcal{L}_{CycleGAN-TTL}$, given by: 

\begin{equation}\label{eq:14}
  \begin{aligned}
  \mathcal{L}_{CycleGAN-TTL} = \mathcal{L}_{CycleGAN} +\lambda_{CE}* \mathcal{L}_{CE-Source} \\+ \lambda_{CE}*\mathcal{L}_{CE-Target},
  \end{aligned}
\end{equation}
where $\mathcal{L}_{CE-Source}$ and $\mathcal{L}_{CE-Target}$ are the cross-entropy losses for the source and target domain classifiers, respectively. 

\subsection{Hybrid CUT (H-CUT) Network}
\begin{figure*} [t]
\begin{center}
\begin{tabular}{c} 
\includegraphics[scale = 0.5]{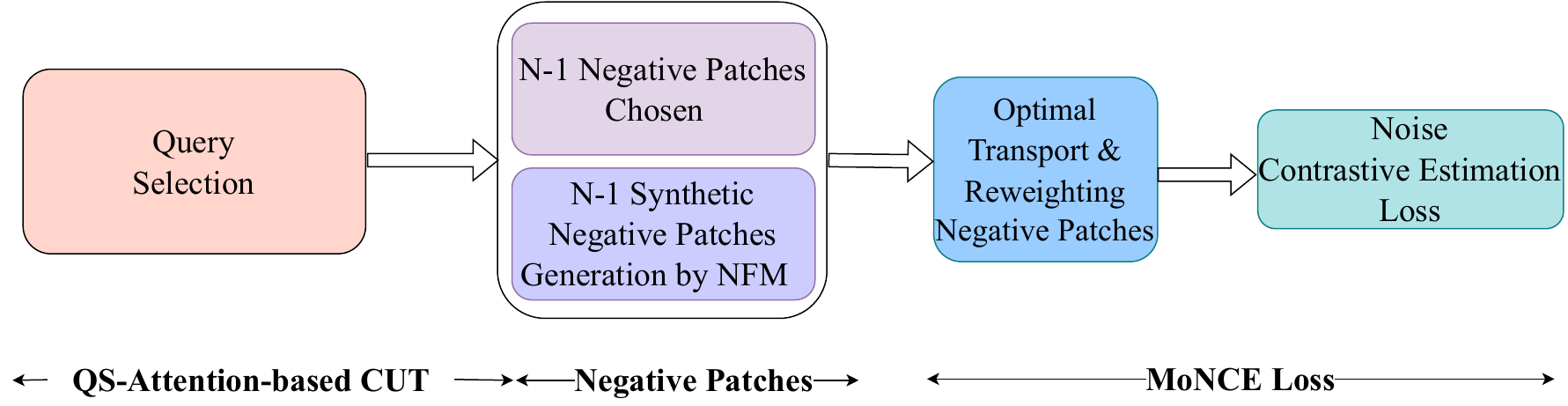}
\end{tabular}
\end{center}
\caption[example] 
{ \label{fig:hcut} 
Block diagram of the proposed H-CUT network incorporating QS-Attention for query selection. A total of N-1 negative patches are chosen from the source domain image, and N-1 synthetic patches are generated using the noisy feature mixup module. Afterwards, a total of 2N-2 patches undergo an optimal transport plan and NCE loss within the MoNCE module.}
\end{figure*}

The proposed H-CUT network is the building block for the contrastive learning-based TTL network (C3TTL). It consists of a contrastive unpaired I2I translation network~\cite{cut} that optimizes the modulated NCE loss~\cite{cut_monce}. Furthermore, the H-CUT network has three components: i) query-selected attention~\cite{cut_qs_attn}, ii) noisy synthetic negative patch generation~\cite{noisy_feature_mixup}, and iii) MoNCE~\cite{cut_monce} module. The components of the H-CUT network are depicted in \figurename ~\ref{fig:hcut}. In this network, both the original and synthetic patches undergo an optimal transport plan, which modulates the hardness of negative patches. It is worth noting that incorporating the NFM module into the H-CUT network provides 2*(N-1) negative patches in the MoNCE loss mentioned in Eq.~\ref{eq:6_patch}.

\subsection{Full Objective}
\begin{figure} [h]
\begin{center}
\begin{tabular}{c} 
\includegraphics[scale = 0.45]{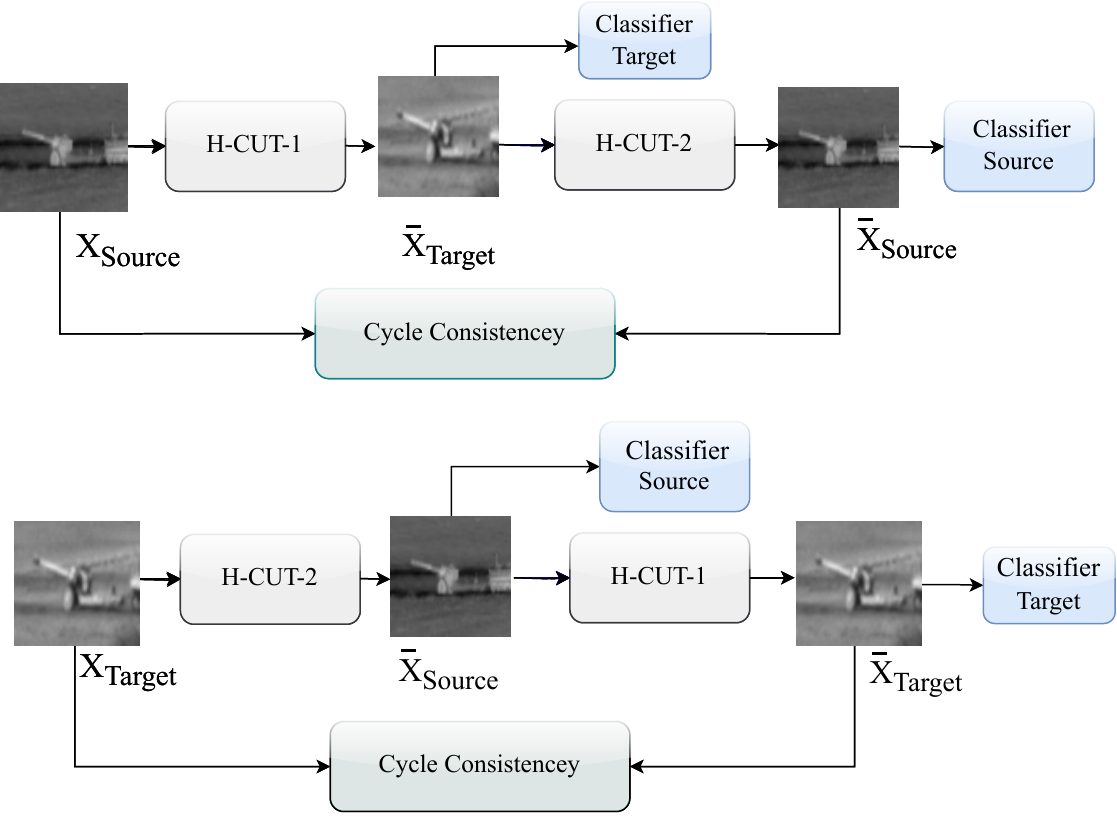}
\end{tabular}
\end{center}
\caption[example] 
{ \label{fig:ttl_cut} 
Block diagram of the C3TTL network, where, H-CUT-1 transforms source domain to target domain images, H-CUT-2 performs the opposite way, cycle-consistency-based bijection mapping preserves the geometric structure of reconstructed and input image, source domain classifier guides the gradually learning target domain classifier. Identity loss is omitted here. }
\end{figure}

The proposed C3TTL network consists of two H-CUT networks for bidirectional image translation between source and target domains. The reconstructed source images are used to compute the categorical-cross-entropy loss by the source domain classifier, which is backpropagated to guide the network to produce class-preserving image generation. Additionally, this TTL network constructs a target domain classifier that will annotate unlabeled target domain images. All components of this network are depicted briefly in the \figurename~\ref{fig:ttl_cut}. The detailed overview of the C3TTL is illustrated in \figurename~\ref{fig:C3TTL}. The architecture, weight, and losses of the source and target domain classifiers are chosen in a similar manner to the CycleGAN-TTL framework. The overall objective of the proposed C3TTL network is a weighted combination of adversarial, MoNCE, cycle-consistency, identity, and categorical cross-entropy losses. The complete objective is as follows:
\begin{equation}\label{eq:15}
\small
\begin{split}
\mathcal{L}_{C3TTL} = \lambda_{1}* \mathcal{L}_{GAN} + \lambda_{2}* \mathcal{L}_{MoNCE_{1}}(G_{XY},D_Y,X) \\ + \lambda_{3}* \mathcal{L}_{MoNCE_{2}} (G_{YX},D_X,\bar{Y}) +\lambda_{4}* \mathcal{L}_{MoNCE_{2}}(G_{YX},D_X,Y) \\ + \lambda_{5}* \mathcal{L}_{MoNCE_{1}}(G_{XY},D_Y,\bar{X}) + \lambda_{6}* \mathcal{L}_{Cycle} + \lambda_{7}*\mathcal{L}_{Identity}+ \\ \lambda_{CE}* (\mathcal{L}_{CE-Source} + \mathcal{L}_{CE-Target}).
\end{split}
\end{equation}

\section{Experiments}\label{sec:experiments}
\subsection{Dataset}
We employ three ATR datasets to evaluate the annotation performance of the C3TTL framework. These three datasets comprise two ATR vehicle datasets and one ship target dataset. All three datasets consist of unpaired ATR source and target domain images that are captured by different sensors under various atmospheric conditions.

\subsubsection{DSIAC dataset}
We implement our TTL approach using the publicly available DSIAC dataset~\cite{dsiac}. This dataset was collected by the US Army Night Vision and Electronic Sensors Directorate (NVESD). The DSIAC dataset contains a total of ten vehicles in the visible and mid-wave infrared (MWIR) domain. Among these classes, two of them are civilian vehicles (\emph{Pickup}, \emph{Sport vehicle}), one is an artillery piece (\emph{D20}), and the rest of the seven are military vehicles (\emph{2S3, BTR70, BRDM2, BMP2, MT-LB, T72, ZSU23-4}). The distance between target vehicles and the camera varies from 1 to 5 kilometers. The dataset contains 189 video sequences in the MWIR domain and 97 video sequences in the visible domain. Each video sequence consists of 1,800 video frames. Generally, the image size in the DSIAC dataset is 640x480 pixels. In this work, we detect and crop the ATR vehicles (target chips) from the DSIAC dataset using the information from the Meta-UDA~\cite{meta_domenic}. All the target vehicles at different ranges are projected into a canonical distance (e.g., 2 kilometers) using bi-cubic interpolation. The final target chip size is 68x68x3. In the DSIAC dataset, we consider the MWIR and visible as the source and the target domain, respectively.

\subsubsection{Visible and Infrared Spectrums (VAIS) Dataset}
The VAIS~\cite{VAIS_dataset} dataset is the world's first publicly available maritime dataset. This dataset comprises 1,623 visible and 1,242 infrared  domain images of maritime vehicles. This dataset has a total of 1,088 corresponding pair images. This dataset has six classes: \emph{Merchant Ship, Medium Passenger Ship, Sailing Ship, Small Boat, Tugboat, and Medium Other Ship}. The distribution of the dataset is depicted in Table~\ref{tab:vais_distribution}. In this experiment, the targets of the VAIS dataset are resized to 68x68x3. Here, we assume the visible and infrared as source and target domains, respectively.

\begin{table}[ht]
\small
\caption{Distribution of classes in the VAIS dataset.} 
\label{tab:vais_distribution}
\begin{center}       
\begin{tabular}{llll} 
\hline
Class Name &Subclass& Number of  & Number of \\
&&IR target& visible target\\\hline

Merchant&Barges&33&35\\
&Cargo&147&149\\\hline
Medium &Ferries&40&42\\
Passenger Ship&Tour Boats&89&98\\\hline
Sailing Ship &Sail Up&159&278\\
&Sail Down&140&134\\\hline
Small Boat &Speed Boat&132&243\\
&Jet-skis&23&58\\
&Small Pleasure Boat&75&82\\
&Large Pleasure Boats&49&65\\

\hline
Tug Boat &&99&57\\\hline
Medium Other &Fishing&28&33\\
Ship&Medium Other&124&152\\\hline
Total&&1138&1416\\\hline

\end{tabular}
\end{center}
\end{table} 
\subsubsection{FLIR ATR Dataset}
For evaluating the proposed TTL methods, we use 12-bit gray-scale mid-wave (MW) and long-wave (LW) FLIR ATR images that are captured by an experimental laboratory infrared sensor~\cite{asif_mehmood1}. Furthermore, a quantum well infrared photodetector focal plane array is used for the LW sensor. In contrast, an indium antimonide (InSb) focal plane array is employed for the MW sensor. The size of the input image is 304x504 pixels. The dataset consists of 461 paired MW and LW infrared images with 572 target chips. The target chips are captured at different poses, with capturing distances ranging from 1 to 4 kilometers.  The target chips are cropped and reshaped into 68x68x3 pixels. The dataset has a total of seven classes of ATR vehicles: \emph{M60A3, PICKUP, HMMWV, M2, M35, M113, and UNKNOWN}. In this dataset, we consider the mid-wave infrared domain images as the source domain and the long-wave infrared domain images as the target domain.

\subsection{Training Details}
The learning rate of the C3TTL network is set to $2\mathrm{e}{-4}$ for the first 30 epochs; afterward, it is reduced to $1\mathrm{e}{-4}$ for the rest of the 20 epochs. For fast convergence of the network, we have initialized the weights of the H-CUT generator from the weights of the CycleGAN trained for the task of summer to winter translation on the Yosemite dataset~\cite{summer_winter}. The hyperparameters of Eq.~\ref{eq:13} and \ref{eq:15} are set to $\lambda_{a}$ = $\lambda_{1}$ = $\lambda_{2}$ = $\lambda_{3}$ = $\lambda_{4}$ = $\lambda_{5}$ = 1, $\lambda_{b}$ = $\lambda_{6}$ = 10, and $\lambda_{c}$ = $\lambda_{7}$ = 5. During training, for the initial 10 epochs, we set the value of $\lambda_{CE}$ = 0.5 in Eq.~\ref{eq:14} and~\ref{eq:15}, after that $\lambda_{CE}$ = 2.5. To optimize the weights of the generators and the discriminators, we use the Adam optimizer~\cite{adam} with $\beta_1$ = 0.5 and $\beta_2$ = 0.999. To stabilize the GAN training, we update the discriminator loss five times less than the generator loss. For training the classifiers, we also use the Adam optimizer with a learning rate of $5\mathrm{e}{-4}$ and $\beta_1$ = 0.9, $\beta_2$ = 0.999 for 40 epochs. We finetune the target domain classifier for 10 epochs to get the result from 1\% and 10\% labeled data as shown in Table~\ref{tab:different_ttl_dsiac}. The other variants of TTL hyperparameters are the same as the C3TTL. These experiments were conducted on an NVIDIA RTX-8000 GPU using the PyTorch framework~\cite{pytorch} with a batch size of 160.

\subsection{Evaluation Metrics}
To quantitatively evaluate the annotation performance of the C3TTL network, we utilize the classification accuracy among unlabeled and partially labeled target domain data. Furthermore, Fr\'echet Inception Distance (FID) \cite{fid} is exploited in the proposed transductive transfer learning network to quantitatively measure the synthetic images without the need for human participation.

\subsection{Ablation Study}
\begin{table*}[ht]
\caption{Performance of different variants of the TTL network on the DSIAC dataset.} 
\label{tab:different_ttl_dsiac}
\begin{center}       
\begin{tabular}{llll} 
\hline
Method Name  & Accuracy (\%)  & Accuracy & Accuracy\\
&(No label data)&(1\% label data)&(10\% label data)\\\hline
CycleGAN TTL&71.56&80.24&94.86\\ 
QS-Attn+PatchNCE TTL&64.87&72.18&87.00\\ 
PatchNCE+cycle-consistency TTL &65.23&81.78&92.87 \\ 
QS-Attn+MoNCE+cycle-consistency C3TTL&76.22&88.40&94.40\\ 
QS-Attn+MoNCE+NMF TTL&70.09&84.74&93.98\\ 
QS-Attn+MoNCE+cycle-consistency+NMF C3TTL&\textbf{76.61}&\textbf{88.49}&\textbf{97.13}\\
SimCLR~\cite{simclr} &---&75.69&95.15\\
BYOL~\cite{byol} &---&78.62&94.43\\
SwAV~\cite{swav} &---&75.42&96.02\\
B-Twins~\cite{btwin} &---&75.50&94.17\\
\hline
\multicolumn{1}{c}{100\% labeled data} &  \\ \hline
ResNet-18 (supervised) &99.28&&\\ \hline

\end{tabular}
\end{center}
\end{table*}

\begin{table}[ht]
\caption{FID score of different variants of the TTL network on the DSIAC dataset.} 
\label{tab:FID_table}
\begin{center}       
\begin{tabular}{llll} 
\hline
TTL Method Name  & FID score \\
\hline
\centering
CycleGAN &216.109\\ 
QS-Attn+PatchNCE &209.372\\ 
PatchNCE+cycle-consistency &190.316\\ 
QS-Attn+MoNCE+NMF &182.686\\ 
QS-Attn+PatchNCE+cycle-consistency  &171.485\\ 

QS-Attn+MoNCE+cycle-consistency (C3TTL)&94.989\\ 

QS-Attn+MoNCE+cycle-consistency+NMF (C3TTL) &\textbf{87.957}\\ \hline
\end{tabular}
\end{center}
\end{table} 
We initially conducted extensive ablation studies to assess the efficacy of the proposed C3TTL network. For this purpose, we evaluate the effectiveness of each component, i.e., cycle-consistency, MoNCE,  QS-Attention, and NFM module, of the C3TTL network. So, the classification accuracy and FID scores of different variants of the TTL network, constructed by removing these components, are depicted in the Table~\ref{tab:different_ttl_dsiac}, Table~\ref{tab:FID_table}, respectively. This ablation study is conducted on the DSIAC dataset.\\

\begin{figure*}[h!]
    \centering
   
    \begin{subfigure}[t]{0.5\textwidth}
        \centering
        \includegraphics[scale =.32]{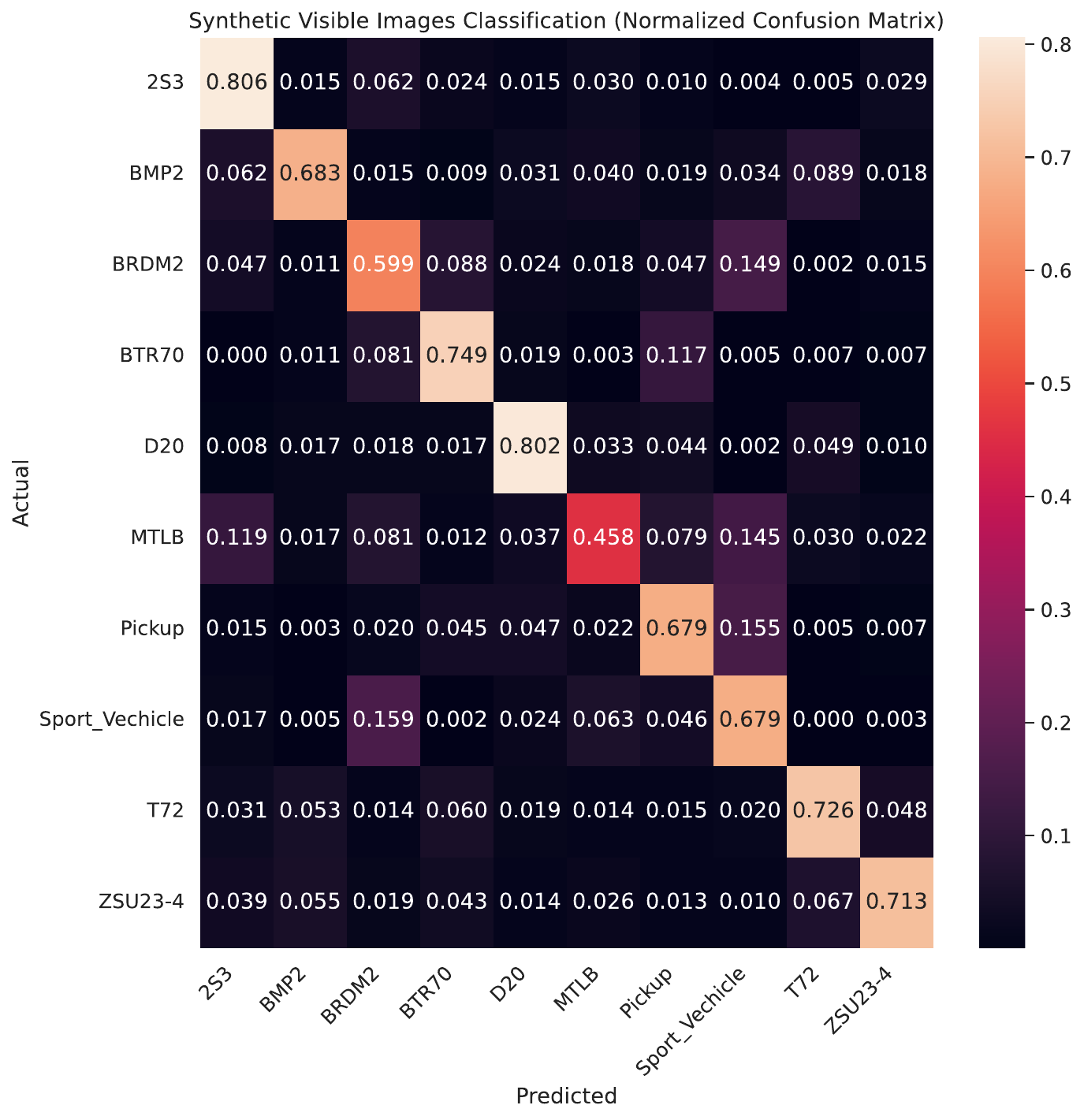}
        \caption[example] 
        {\label{fig:ttl_cut_8a}
        Confusion Matrix (QS-Attn+PatchNCE based TTL network (no cycle-consistency)).}

    \end{subfigure}%
    ~ 
    \begin{subfigure}[t]{0.5\textwidth}
        \centering
        \includegraphics[scale=0.32]{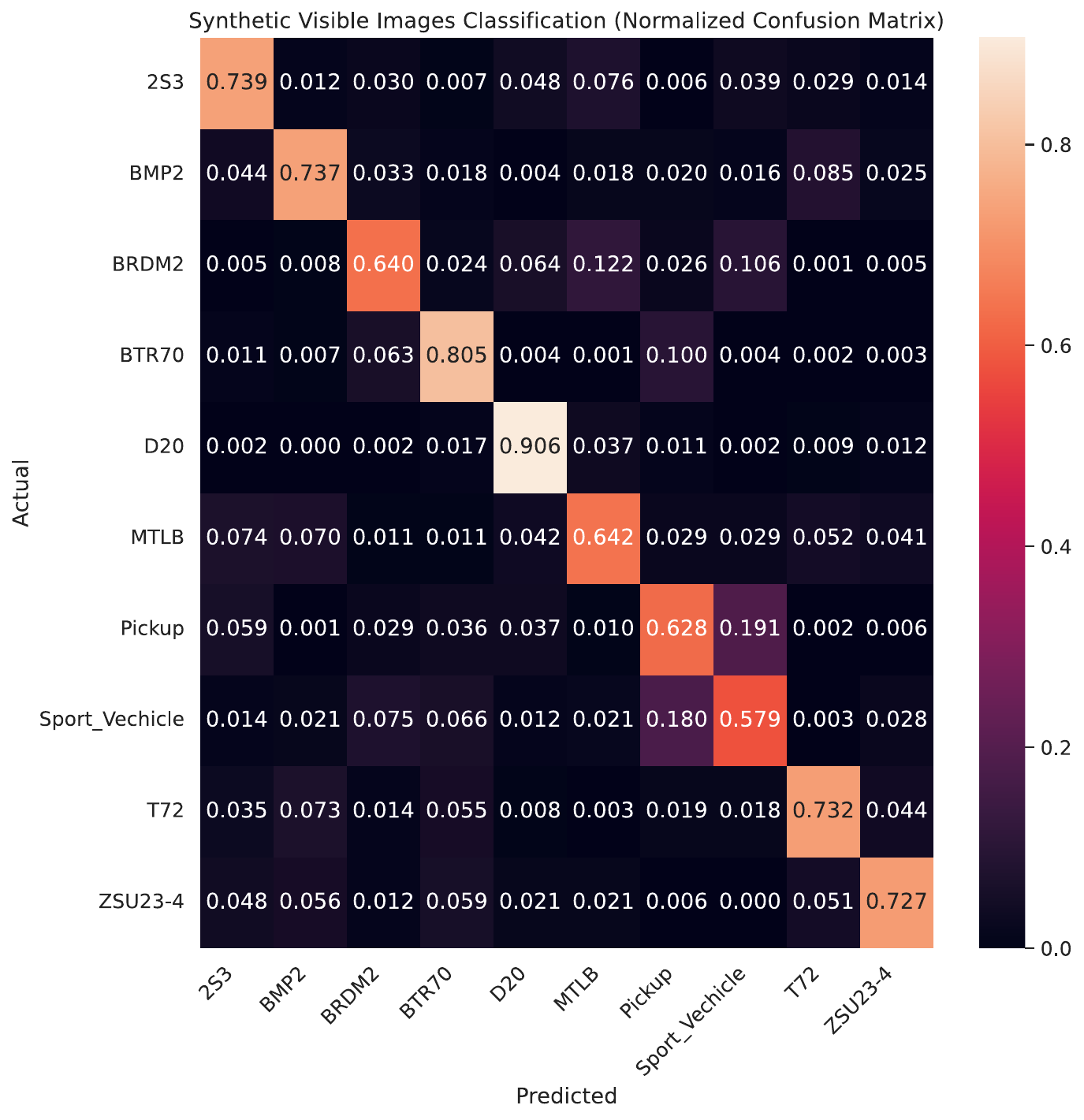}
        \caption{\label{fig:ttl_cut_8b}
        Confusion Matrix (QS-Attn+MoNCE+NFM  based TTL network (no cycle-consistency)).}
    \end{subfigure}
        \centering
    \begin{subfigure}[t]{0.5\textwidth}
        \centering
        \includegraphics[scale =.32]{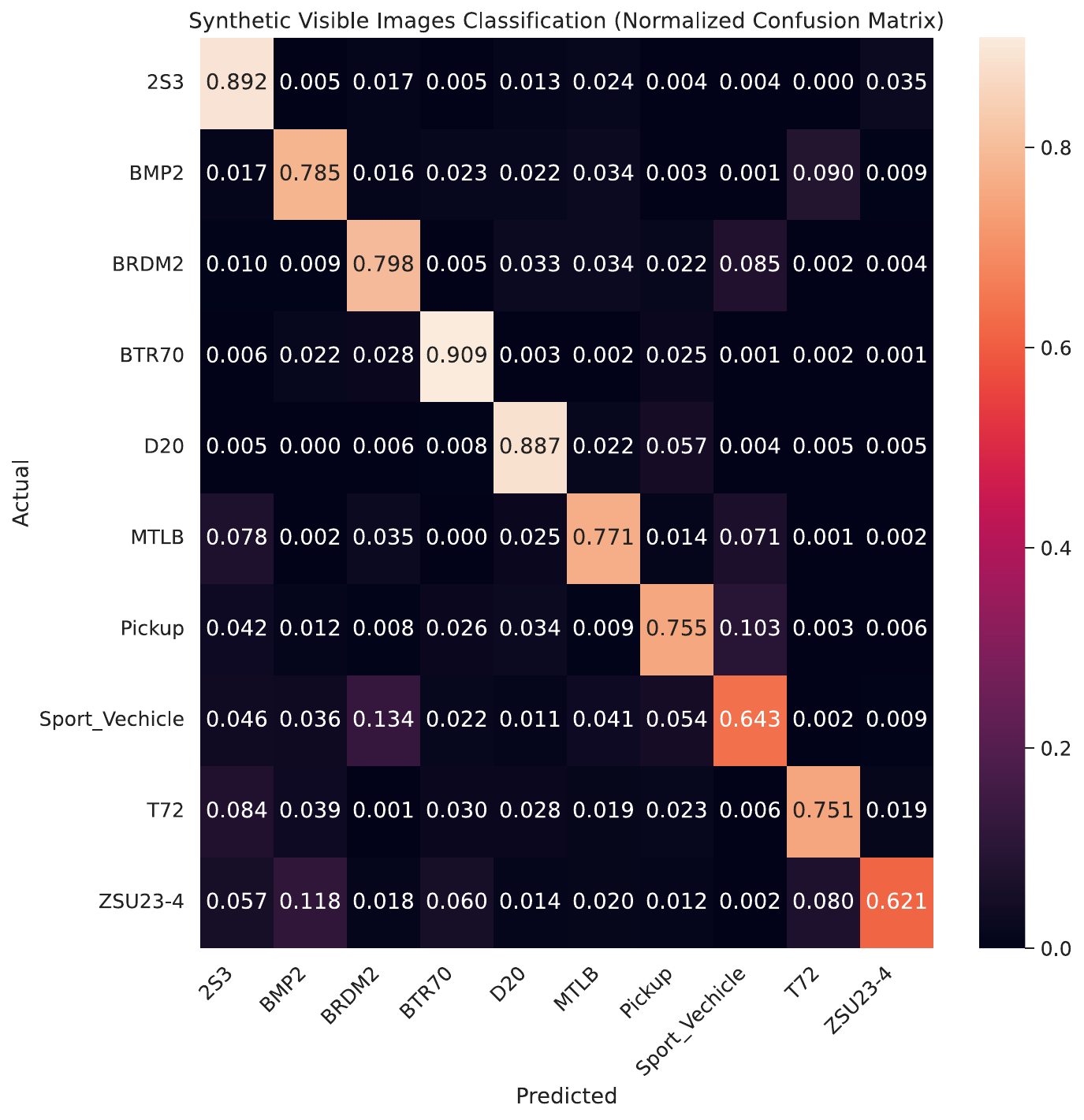}
        \caption[example] 
        {\label{fig:ttl_cut_8c}
        Confusion Matrix (QS-Attn+MoNCE+cycle-consistency based TTL network).}

    \end{subfigure}%
    ~ 
    \begin{subfigure}[t]{0.5\textwidth}
        \centering
        \includegraphics[scale=0.32]{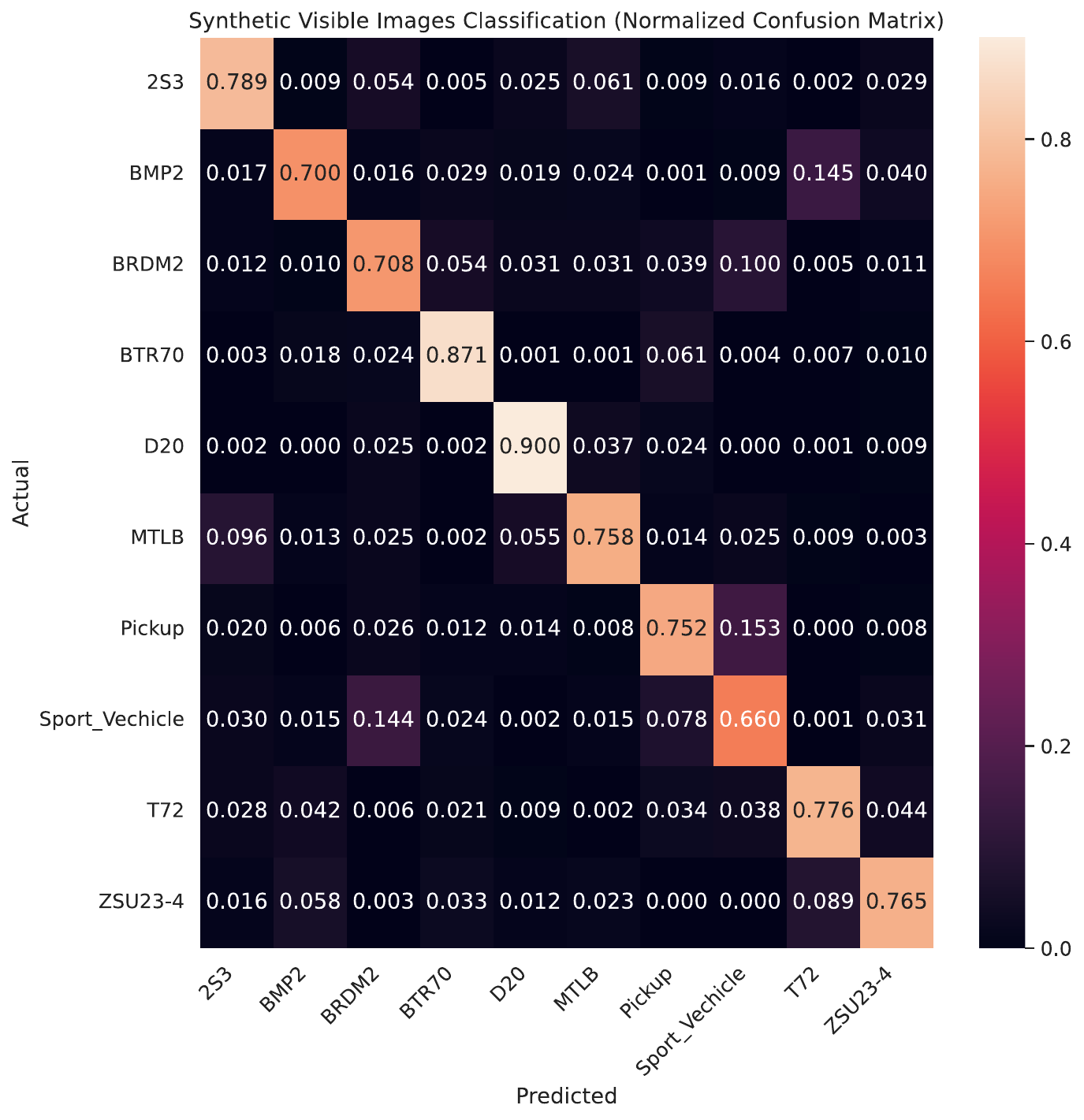}
        \caption{\label{fig:ttl_cut_8d}
        Confusion Matrix (QS-Attn+MoNCE+NFM+cycle-consistency based TTL network).}
    \end{subfigure}

    \caption{Confusion matrices of the target domain classifiers by the C3TTL network using the DSIAC dataset (no target domain label data)).}
\end{figure*}

\textbf{The  effect  of  cycle-consistency.} 
The C3TTL method employs two H-CUT networks to achieve a two-way translation while preserving cycle-consistency. In our ablation experiments, we first eliminate the cycle-consistency loss and analyze the impact on annotation performance and the FID score change in the DSIAC dataset. From \figurename ~\ref{fig:ttl_cut_8a} and \figurename ~\ref{fig:ttl_cut_8c}, we can deduce that cycle-consistency significantly improves the annotation performance in eight out of ten classes, with the exception of \emph{Sports Vehicle} and \emph{ZSU23-4}. Furthermore, cycle-consistency in \figurename ~\ref{fig:ttl_cut_8b} and \figurename ~\ref{fig:ttl_cut_8d} improves the average annotation performance in C3TTL by 9.3\%; however, exceptions are found in the \emph{BMP2} and \emph{D20} classes. Therefore, cycle-consistency is essential in the C3TTL network because it preserves the geometric structure and guides the whole TTL network bijectively. 

\textbf{The  effect  of  MoNCE.} In the C3TTL, we also investigate the benefits of the MoNCE~\cite{cut_monce} loss over PatchNCE~\cite{cut} loss in Table~\ref{tab:different_ttl_dsiac} and Table~\ref{tab:FID_table}. From this observation, we can infer that MoNCE significantly improves the performance of the TTL network as MoNCE loss help to converge the H-CUT network more effectively than vanilla PatchNCE loss.

\textbf{The  effect  of  QS-Attention.} We also investigate the necessity of the QS-Attention~\cite{cut_qs_attn} module in the C3TTL network. Table~\ref{tab:FID_table} shows that the QS-Attention module improves synthetic image quality in the C3TTL network, especially in the PatchNCE+cycle-consistency-based TTL. Adding QS-Attention with it improves the FID score from 190.32 to 171.49. The reason is that in the CUT network, the QS-Attention intentionally selects a query from the region relevant to the domain. Thus, it improves the unpaired I2I system as well as the C3TTL network.

\textbf{The  effect  of  synthetic patches by NFM module.}  We analyze the necessity of synthetic negative patches introduced by the noisy feature mixup~\cite{noisy_feature_mixup} module in the TTL network and occasionally observe improved performance. From Table~\ref{tab:different_ttl_dsiac}, we observe that the incorporation of the negative patches produces a slight improvement (0.51\%) in the performance of the C3TTL network for the DSIAC dataset; however, it decreases the performance slightly for the VAIS and FLIR ATR datasets. From \figurename ~\ref{fig:ttl_cut_8c} and \figurename ~\ref{fig:ttl_cut_8d}, the NFM module improves the annotation performance of the TTL network in a total of four classes, i.e., \emph{D20}, \emph{Sport Vehicle}, \emph{T72}, and \emph{ZSU23-4}. Moreover, Table~\ref{tab:FID_table} shows that the NMF module of the TTL network improves the FID score ($94.989 \rightarrow 87.957$) in the DSIAC dataset.\\
Generally, the proposed C3TTL network performs best when all four components are utilized together. Furthermore, the NFM module has less impact on the C3TTL network than the cycle-consistency, MoNCE, and QS-Attention modules. The synthetic negative patches could potentially be more effective if the NFM module mixes negative patches with the query or mixes the hardest negative patches, as proposed in \cite{hard_negative_Mochi}. In this work, we only mix the negative patches, not the hardest negative patches. In the future, we will investigate the effect of mixing the hardest negative~\cite{hard_negative_Mochi} patches in the TTL network.

\subsection{Numerical and Visual Evaluation}
In this sub-section, we will evaluate the supervised source and target domain classifiers as well as the target domain classifier of the C3TTL network without using any labeled target domain data. Furthermore, we will compare the performance of our proposed C3TTL method with state-of-the-art approaches.

\begin{figure*}[h!]
    \centering
    \begin{subfigure}[t]{0.4\textwidth}
        \centering
        \includegraphics[scale =.38]{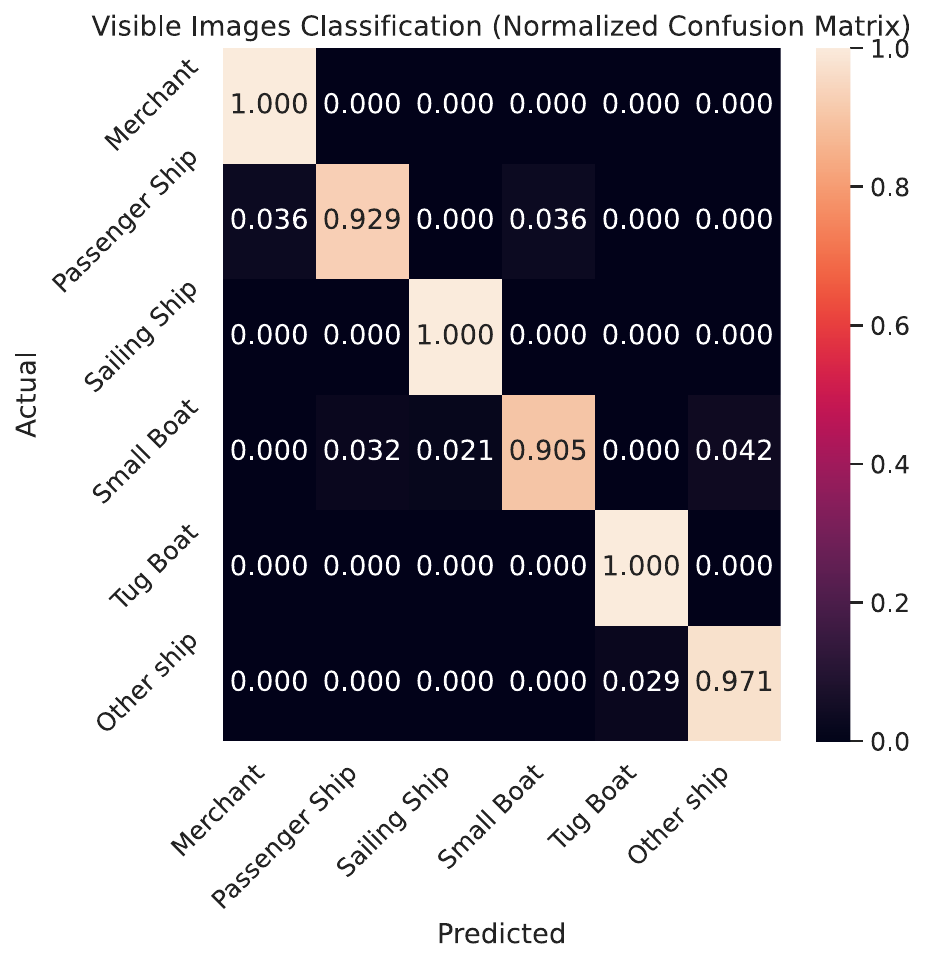}
        \caption{\label{fig:VIS_atr-2-a}
        Visible domain classifier using VAIS dataset (100\% labeled data).}
    \end{subfigure}%
    ~
    \begin{subfigure}[t]{0.4\textwidth}
        \centering
        \includegraphics[scale=0.38]{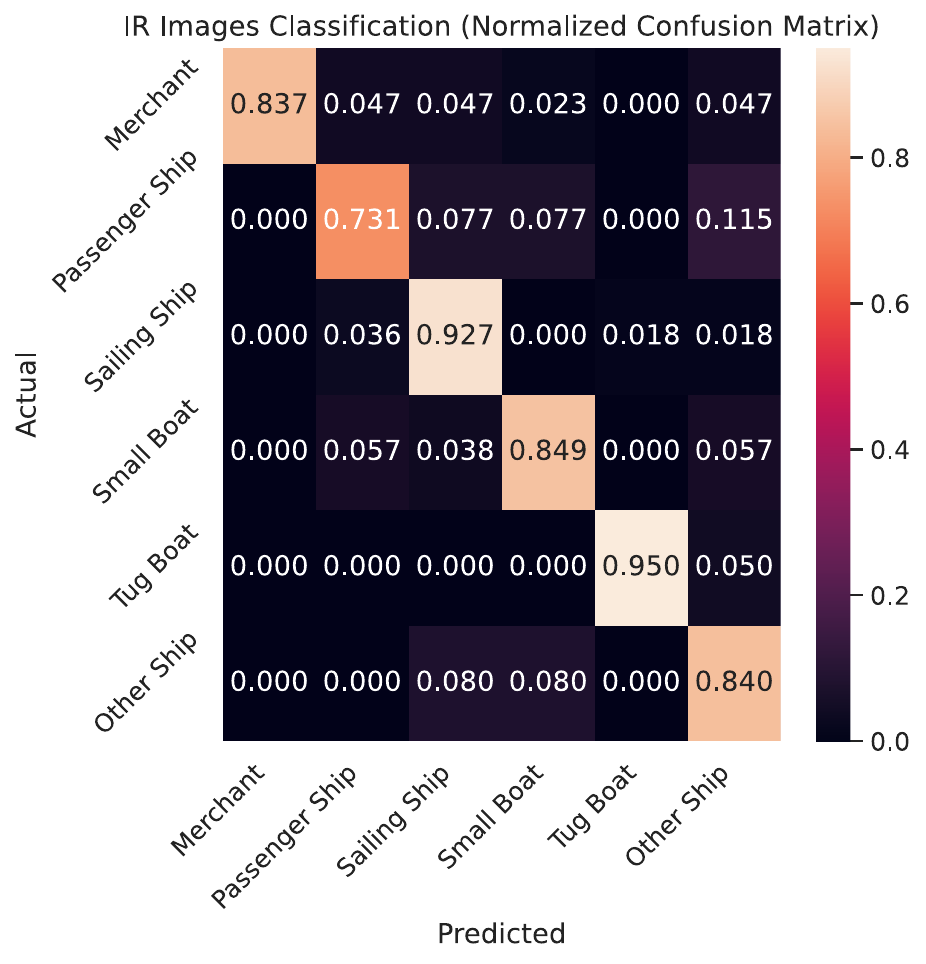}
        \caption{\label{fig:VIS_atr-2-b}
        IR domain classifier using VAIS dataset (100\% labeled data).}
    \end{subfigure}

    \begin{subfigure}[t]{0.4\textwidth}
        \centering
        \includegraphics[scale =.36]{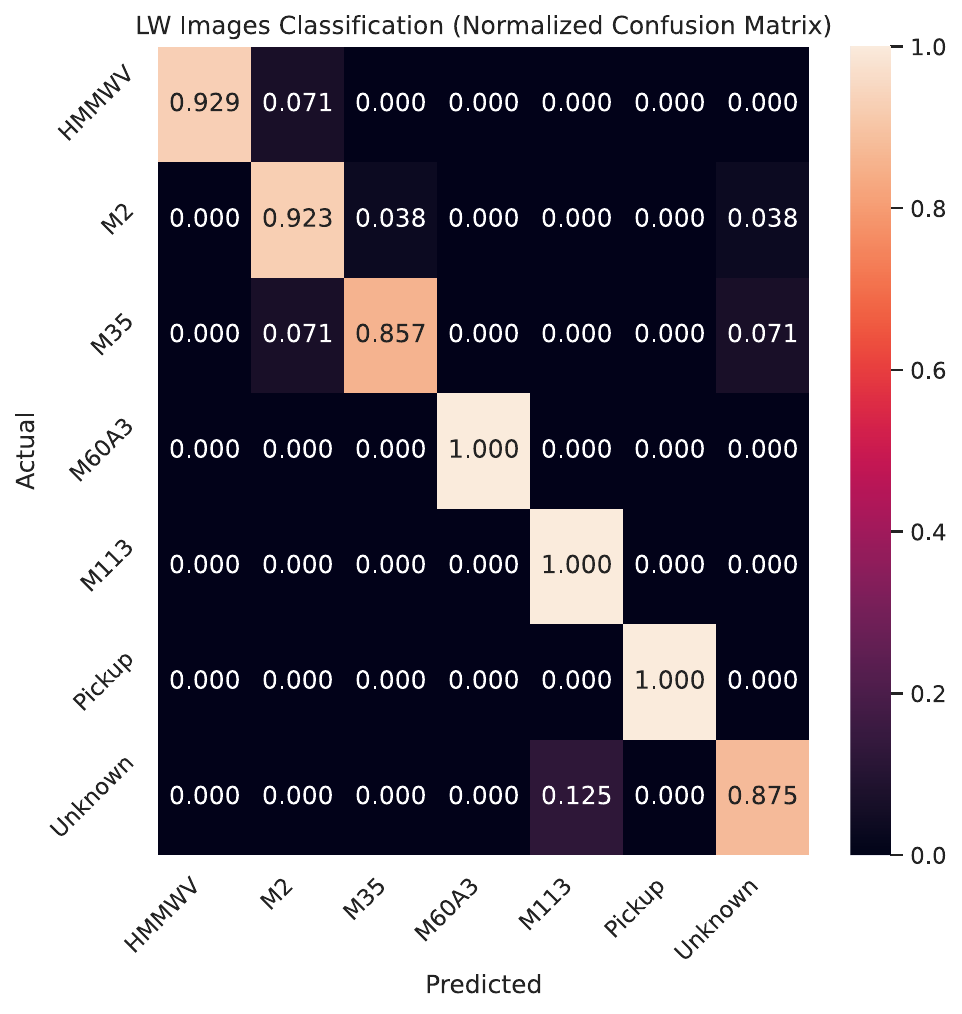}
        \caption{\label{fig:VIS_atr-2-c}
        LW domain classifier using FLIR ATR dataset (100\% labeled data).}
    \end{subfigure}%
    ~
    \begin{subfigure}[t]{0.4\textwidth}
        \centering
        \includegraphics[scale=0.36]{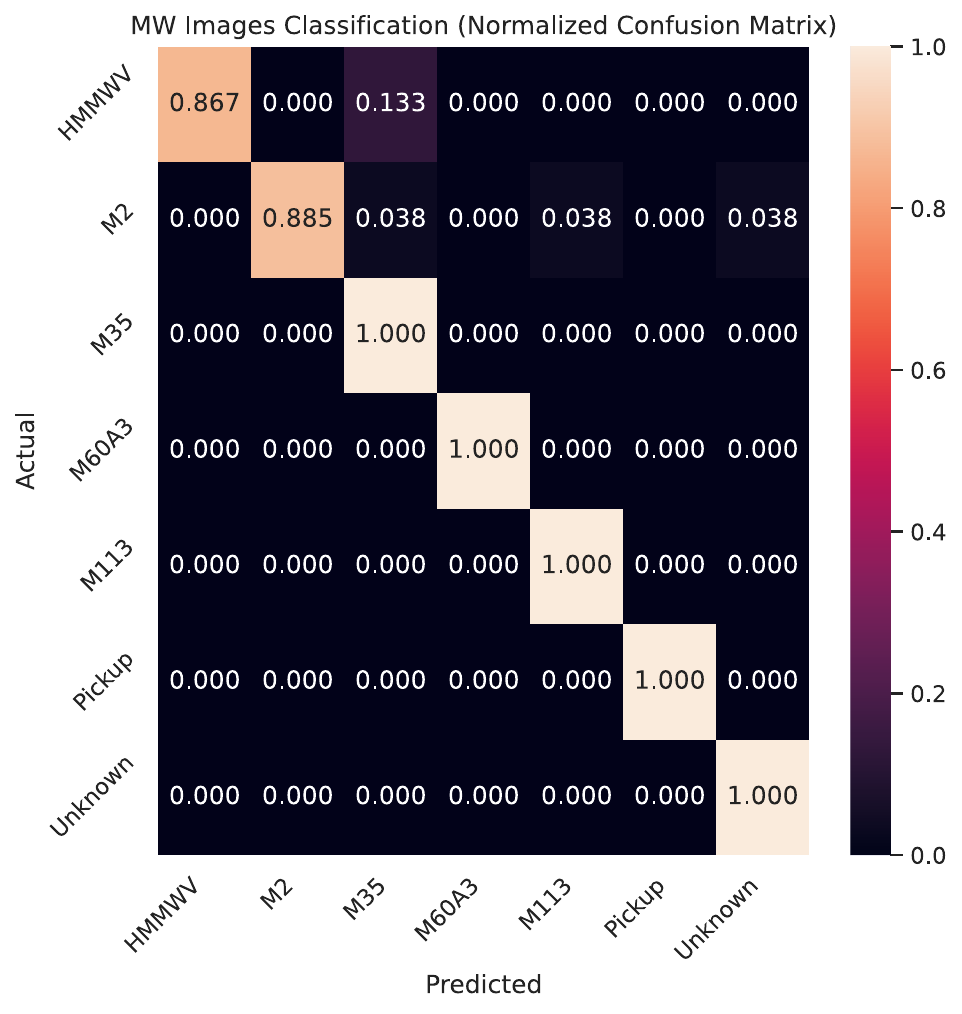}
        \caption{\label{fig:VIS_atr-2-d}
        MW domain classifier using FLIR ATR dataset (100\% labeled data).}
    \end{subfigure}

    \caption{Confusion matrices of the source and target domain classifiers using the VAIS and FLIR ATR datasets. }
\end{figure*}
 \begin{figure*}[h!]

   \includegraphics[scale = 0.32]{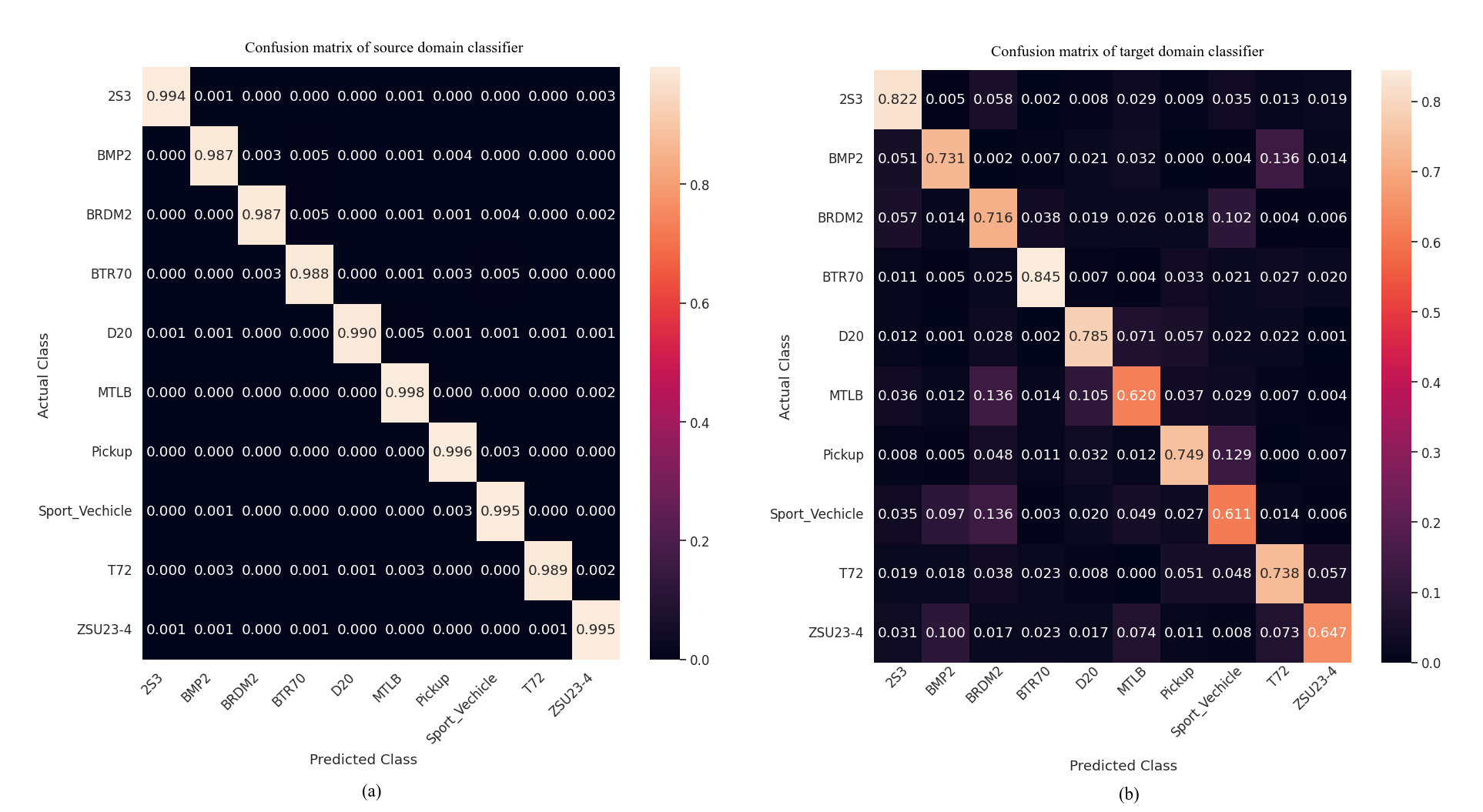}
   
   \captionsetup{justification=centering}
   \caption[example] 
   { \label{fig:example_confusion} 
Confusion matrices of (a) source domain classifier and (b) target domain classifier (no labeled data) using the transductive CycleGAN network in the DSIAC dataset~\cite{TransductiveCycleGAN}.}
   \end{figure*}
\begin{figure}[ht!]
    \includegraphics[scale=0.32]{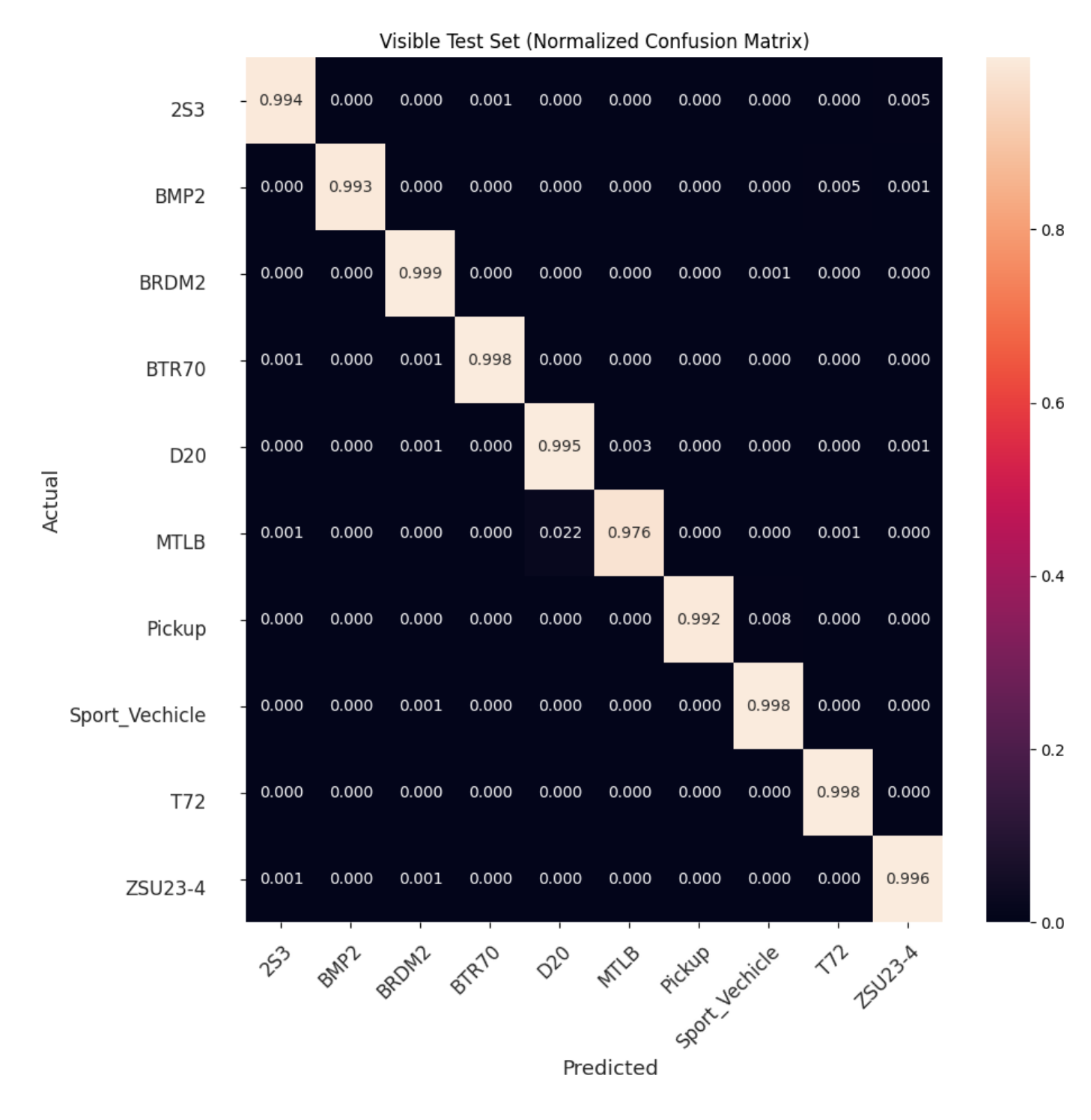}
    \captionsetup{justification=centering}

    \caption{Confusion matrices of supervised target domain classifier (100\% labeled data) in the DSIAC dataset.}
    \label{fig:enter-label}
\end{figure}

\subsubsection{Performance of Supervised Classifier}
\textbf{Results on DSIAC Dataset.}  We evaluate the performance of the source and target domain classifiers using the DSIAC dataset. The training, testing, and validation sets are constructed by randomly dividing the DSIAC dataset into 70:15:15 ratios, respectively.  The confusion matrix of the source and the target domain classifiers are illustrated in \figurename ~\ref{fig:example_confusion}(a), and~\ref{fig:enter-label}. The confusion matrix provides the normalized performance of the classifier. The accuracy of the supervised source and target domain classifies is 99.16\% and 99.28\%, respectively. Additionally, the performance of the source domain classifier with targets at different distances is investigated in Table~\ref{tab:distance_performance}. From this table, we can infer that the classification performance of the source domain classifier is almost consistent when capturing images at distances between 1-4 kilometers. Although, the performance of the source classifier is degraded when the capturing distance of the target exceeds four kilometers. As the capturing distance increases, the image quality of the target chips reduces in comparison to target chips captured at shorter distances.

\begin{table}[ht]
\caption{Performance of the source domain classifier at different target distances in the DSIAC dataset.} 
\label{tab:distance_performance}
\begin{center}       
\begin{tabular}{lll} 
\hline
Target Distance (m)&Accuracy(\%)&Number of Samples\\\hline
1000&99.47&4,885\\
1500&	99.72&	4,713\\
2000&	99.78&	4,895\\
2500&	99.54&	4,830\\
3000&	99.45&	4,558\\
3500&	99.76&	4,193\\
4000&	99.11&	4,506\\
4500&	97.83&	4,196\\
5000&	93.55&	4,106\\\hline
\end{tabular}
\end{center}
\end{table} 

\begin{table}
\caption{Performance of the TTL network on the VAIS dataset.} 
\label{tab:different_ttl_performance_VAIS}
\begin{center}       
\begin{tabular}{ll} 
\hline
Method Name  & Average \\ 
& Accuracy (\%)\\ \hline
\multicolumn{1}{c}{No labeled data} &  \\ \hline
CycleGAN TTL&53.60\\ 

QS-Attn+MoNCE+cycle-consistency C3TTL&\textbf{55.41}\\ 
QS-Attn+MoNCE+cycle-consistency+NMF C3TTL&54.50\\ \hline
\multicolumn{1}{c}{100\% labeled data} &  \\ \hline
ResNet-18 (supervised) &86.04\\ \hline

\end{tabular}
\end{center}
\end{table}

\begin{table}
\caption{Performance of the TTL network on the FLIR ATR dataset.} 
\label{tab:different_ttl_perform_ATR-2}
\begin{center}       
\begin{tabular}{ll} 
\hline
Method Name  & Average \\ 
& Accuracy (\%)\\ \hline
\multicolumn{1}{c}{No labeled data} &  \\ \hline
CycleGAN TTL&78.95\\ 

QS-Attn+MoNCE+cycle-consistency C3TTL&\textbf{80.00}\\ 
QS-Attn+MoNCE+cycle-consistency+NMF C3TTL&78.95\\ \hline
\multicolumn{1}{c}{100\% labeled data} &  \\ \hline
ResNet-18 (supervised) &93.68 \\ \hline

\end{tabular}
\end{center}
\end{table} 
 
\begin{figure*} [h!]
    
   \begin{center}
   \begin{tabular}{c} 
   \includegraphics[scale = 0.62]{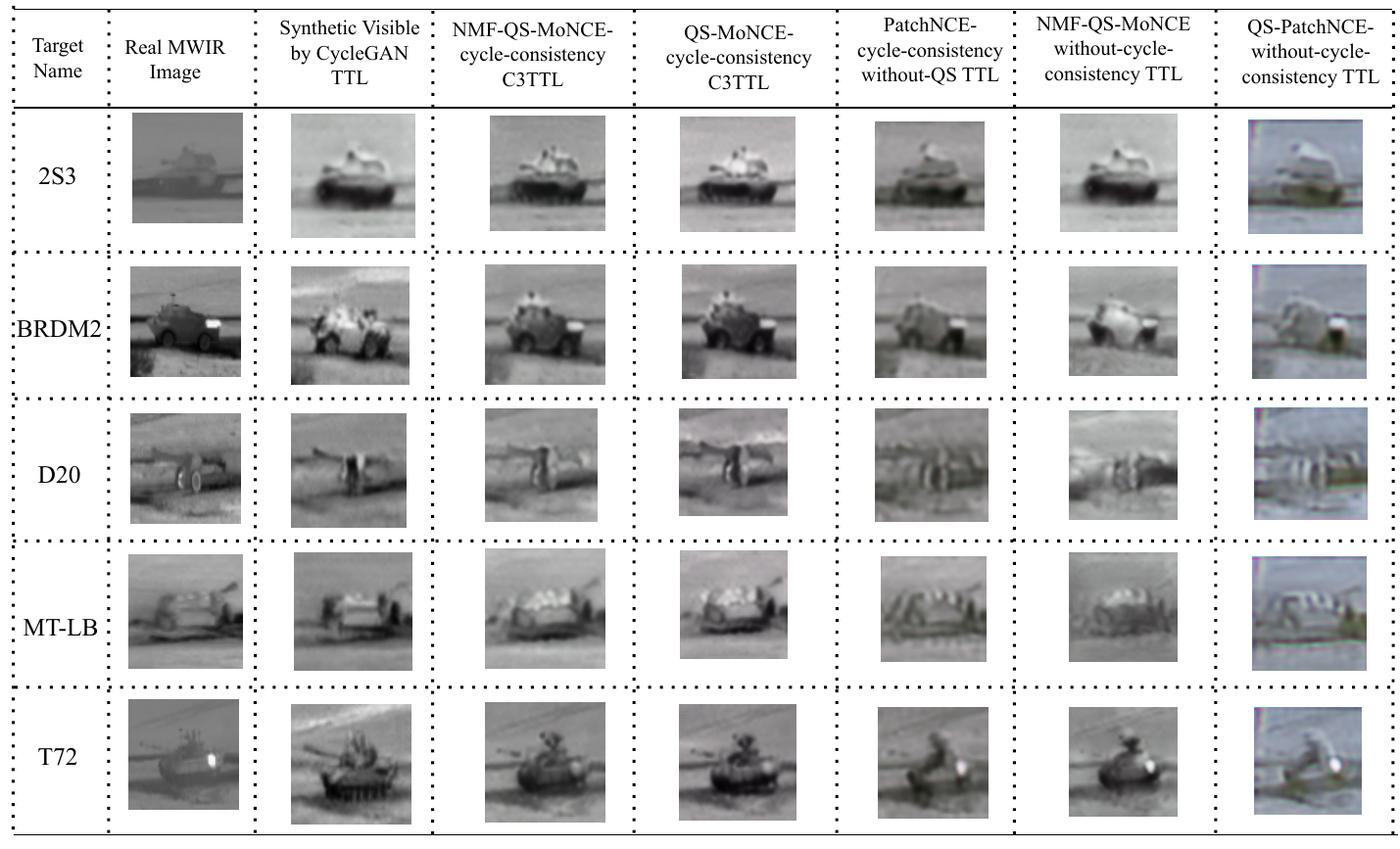}
   \end{tabular}
   \end{center}
   \caption[example] 
   { \label{fig:synthetic_images_transductive} 
Generated images by the proposed TTL framework using the DSIAC dataset.}
\end{figure*}

\begin{figure*} [h!]
   \begin{center}
   \begin{tabular}{c} 
   \includegraphics[scale = 0.7]{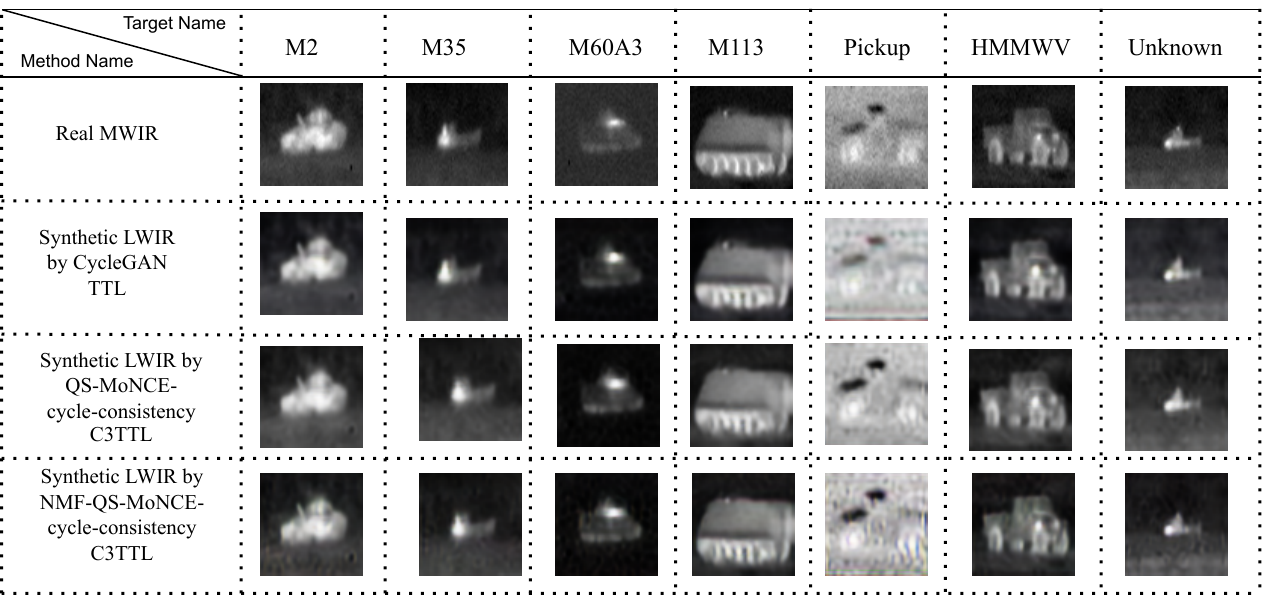}
   \end{tabular}
   \end{center}
   \caption[example] 
   { \label{fig:synthetic_images_transductive_flir_atr} 
Generated images by the proposed TTL framework using the FLIR ATR dataset.}
\end{figure*}

\textbf{Results on VAIS Dataset.}   We further investigate the classification performance of visible (source) and infrared (target) images on the VAIS ship target dataset~\cite{VAIS_dataset}. This dataset is divided into train, test, and validation using a 70:20:10 ratio, respectively. The confusion matrices of the supervised visible and infrared classifiers for the VAIS dataset are depicted in \figurename ~\ref{fig:VIS_atr-2-a} and \figurename ~\ref{fig:VIS_atr-2-b}, respectively. The accuracy of visible and IR domain classifiers is 96.73\% and 86.04\%, respectively in the supervised settings.

\textbf{Results on FLIR ATR Dataset.} Furthermore, we analyze the performance of mid-wave (source domain) and long-wave infrared (target domain) ATR target classification for the FLIR ATR dataset.~\cite{asif_mehmood1}. The train, test, and validation ratio is 70:20:10. The confusion matrices of the supervised source and target domain classifiers are illustrated in \figurename~\ref{fig:VIS_atr-2-c} and \figurename ~\ref{fig:VIS_atr-2-d}, respectively. The classification accuracy of the source and target domain classifier is 95.95\%, and 93.68\%, respectively, in the supervised setting.

\subsubsection{Annotation Performance of TTL Network}
Here, we evaluate the performance of the CycleGAN-TTL and C3TTL in three ATR datasets. The CycleGAN-TTL was utilized in our conference paper~\cite{TransductiveCycleGAN}, however, the C3TTL is an upgraded version proposed in this article. Both TTL networks can successfully annotate the target domain images without any labeled information from the target domain.\\ 

\textbf{Results on DSIAC Dataset.}  The confusion matrix of the target domain classifier by CycleGAN-TTL in the DSIAC dataset is illustrated in \figurename~\ref{fig:example_confusion}(b). The accuracy of the target domain classifier is 71.6\% for the visible domain images in the DSIAC dataset. We also evaluate the performance of the target domain classifier using a fraction of the labeled data. The accuracy of the target domain classifier is 80.24\%, 90.79\%, and 94.86\%, with 1\%, 5\%, and 10\% of the labeled target domain dataset, respectively. The performance of the target domain classifier with 1\% and 10\% labeled data are depicted in Table~\ref{tab:different_ttl_dsiac}. The C3TTL annotation performance is 76.61\%, which is 7.06\% higher than CycleGAN-TTL. The C3TTL can annotate 88.49\% and 97.13\% accurately, with 1\% and 10\% labeled data in the target domain.\\

\textbf{Results on VAIS Dataset.}   In the VAIS dataset, we evaluate the performance of C3TTL and CycleGAN-TTL which is 55.41\% and 53.60\%, respectively. Table~\ref{tab:different_ttl_performance_VAIS} shows the performance of proposed TTL networks on the VAIS dataset.\\

\textbf{Results on FLIR ATR Dataset.} We further evaluate the classification performance of the C3TTL and CycleGAN-TTL in the FLIR ATR dataset. Table~\ref{tab:different_ttl_perform_ATR-2} delineates the accuracy of proposed TTL networks on the FLIR ATR dataset. We achieve a classification accuracy of 80.00\% and 78.95\% for the C3TTL and CycleGAN-TTL, respectively. Moreover, across all datasets, the C3TTL consistently outperforms the CycleGAN-TTL, as the H-CUT network in the C3TTL can transfer the domain more effectively than the CycleGAN.

\subsubsection{Performance Comparison with State-of-the-Art Methods }
We compare our proposed C3TTL and CycleGAN-TTL with SimCLR~\cite{simclr}, BYOL~\cite{byol}, SwAV~\cite{swav}, and B-Twins~\cite{btwin} in Table~\ref{tab:different_ttl_dsiac}. We observe that the annotation performance of the C3TTL outperforms the SimCLR, BYOL, SwAV, CycleGAN-TTL, and B-Twins in the partially labeled (1\% and 10\%) target domain data on the DSIAC dataset. Furthermore, CycleGAN-TTL outperforms SimCLR, BYOL, SwAV, and B-Twins in the 1\% labeled data. However, in the 10\% labeled data, the annotation performance of CycleGAN-TTL is lower compared to the SimCLR and SwAV. In conclusion, we can infer that the proposed C3TTL method is more potent than several state-of-the-art semi-supervised algorithms.

\subsubsection{Visual Evaluation}
The synthetic images generated by the C3TTL and the CycleGAN-TTL are illustrated in \figurename ~\ref{fig:synthetic_images_transductive}, and~\ref{fig:synthetic_images_transductive_flir_atr} using DSIAC and FLIR ATR dataset. From this figure, we observe that CycleGAN-TTL produces blur images compared to C3TTL. Furthermore, the last three columns of \figurename~\ref{fig:synthetic_images_transductive} illustrate that without cycle-consistency and MoNCE, the synthetic images generated by the TTL networks exhibit artifacts and blurriness.
\\
Furthermore, the FID score of different variants of the TTL methods is illustrated in Table~\ref{tab:FID_table}. This table demonstrates that the combination of MoNCE, QS-Attention, NFM, and cycle-consistency in the C3TTL secures the best FID score (87.96) on the DSIAC dataset. 

\subsubsection{Discussion}
The classification performance of different variants of the TTL network in the DSIAC and FLIR ATR datasets is better than the VAIS dataset. It may be because the VAIS dataset is noisy, and the information gap stays between the visible and the infrared domains. In future work, we will investigate the reasons for the low performance of all the TTL networks in the VAIS dataset.

\section{Conclusion}\label{sec:conclusion}
In this paper, we introduce a novel transductive transfer learning network, called C3TTL, to annotate unlabeled ATR images. We improve the unpaired I2I translation system by simultaneously applying modulated contrastive force to the negative patches, domain-relevant query selection, and synthetic negative patch generation, within the contrastive unpaired image translation network. The proposed hybrid CUT network in the TTL improves the annotation performance in several ATR datasets. Extensive experimental results on the DSIAC dataset demonstrate that the proposed C3TTL can annotate more accurately than numerous state-of-the-art methods.

\section*{ACKNOWLEDGMENT}
 This material is based upon work supported in part by the U. S. Army Research Laboratory and the U. S. Army Research Office under contract number: W911NF2210117.
\bibliography{main}
\bibliographystyle{IEEEtran}
\begin{IEEEbiography}[{\includegraphics[width=1in,height=1.25in,clip,keepaspectratio]{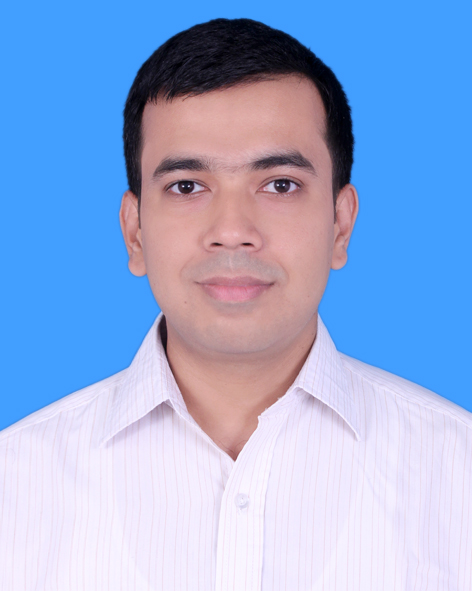}}]{Shoaib Meraj Sami} received the B.Sc. and the M.Sc. degree in Electrical and Electronic Engineering degree from the Bangladesh University of Engineering and Technology, Bangladesh, in 2016 and 2021, respectively. He also worked as an engineer at Nuclear Power Plant Company Bangladesh Limited and Chittagong Port Authority. He is currently pursuing a Ph.D. degree with the Lane Department of Computer Science and Electrical Engineering, West Virginia University, Morgantown, WV, USA. His main focus is on the development of machine learning and deep learning algorithms and their applications with computer vision, biometrics, and automatic target recognition.
\end{IEEEbiography}

\begin{IEEEbiography}[{\includegraphics[width=1in,height=1.25in, clip,keepaspectratio]{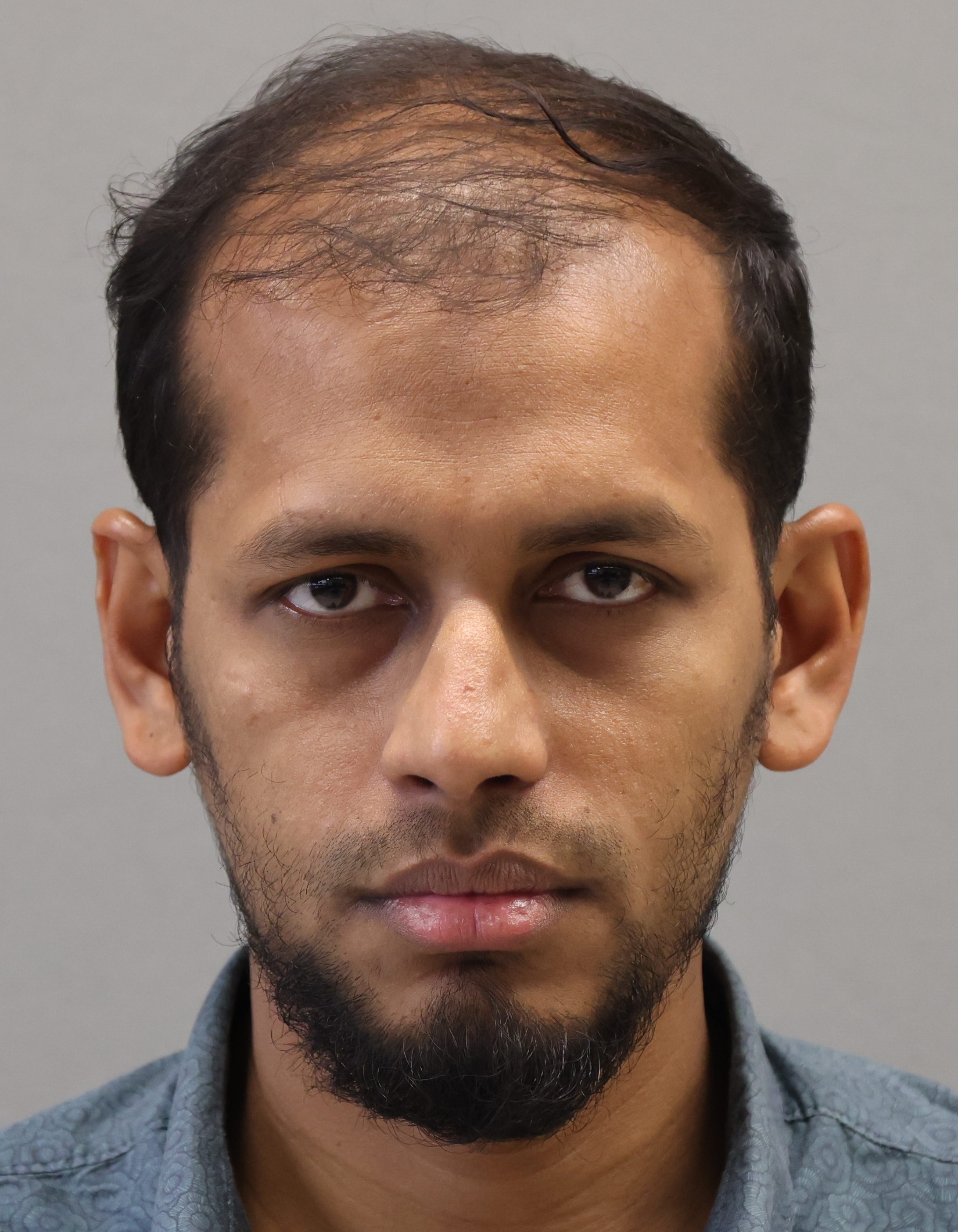}}] {Md Mahedi Hasan} is a third-year PhD student in the Lane Department of Computer Science and Electrical Engineering (LCSEE) at West Virginia University (WVU) in Morgantown, WV, USA. He obtained his B.Sc. in Electrical and Electronic Engineering and M.Sc. in Information and Communication Technology from Bangladesh University of Engineering and Technology (BUET) in 2020. Additionally, he has experience working as a lecturer at Manarat International University (MIU) in Dhaka, Bangladesh. His primary research interests lie in the development of machine learning and deep learning algorithms, focusing on their applications in computer vision and biometrics.
\end{IEEEbiography}
\begin{IEEEbiography}[{\includegraphics[width=1in,height=1.25in,clip,keepaspectratio]{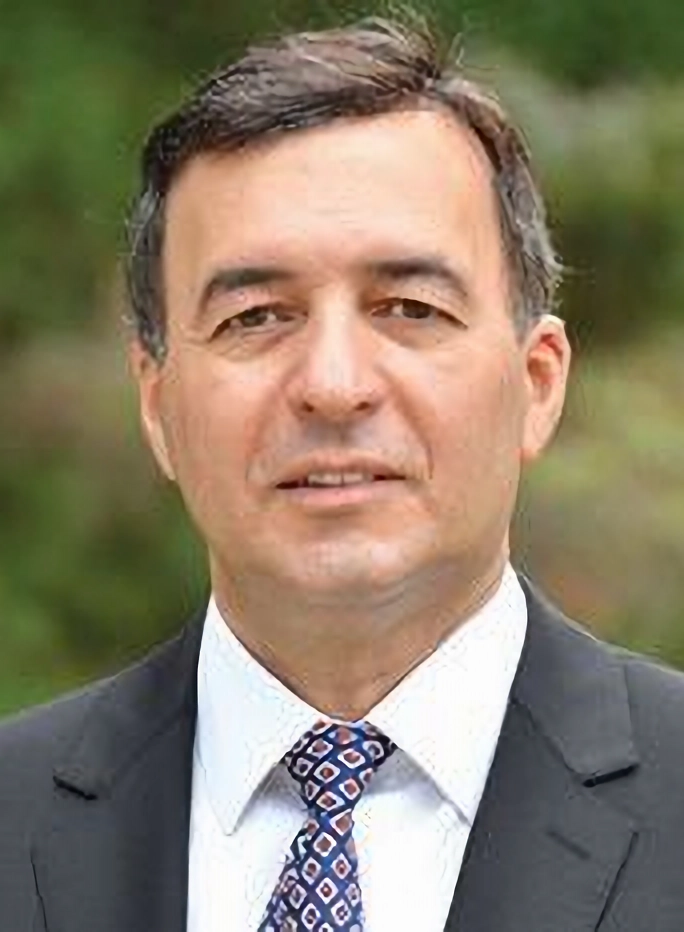}}]{Nasser M. Nasrabadi}(Fellow, IEEE) received the B.Sc. (Eng.) and Ph.D. degrees in
electrical engineering from the Imperial College
of Science and Technology, University of London,
London, U.K., in 1980 and 1984, respectively.
In 1984, he was at IBM, U.K., as a Senior Programmer. From 1985 to 1986, he was at the Philips
Research Laboratory, New York, NY, USA, as a
member of the Technical Staff. From 1986 to 1991,
he was an Assistant Professor at the Department
of Electrical Engineering, Worcester Polytechnic Institute, Worcester, MA,
USA. From 1991 to 1996, he was an Associate Professor at the Department
of Electrical and Computer Engineering, State University of New York at
Buffalo, Buffalo, NY, USA. From 1996 to 2015, he was a Senior Research
Scientist at the U.S. Army Research Laboratory. Since 2015, he has been
a Professor with the Lane Department of Computer Science and Electrical Engineering. His current research interests include image processing,
computer vision, biometrics, statistical machine learning theory, sparsity,
robotics, and deep neural networks. He is a fellow of the \textsc{International Society for Optical
Engineers (SPIE)}. He has served
as an Associate Editor for the \textsc{Ieee Transactions On Image Processing}, the
\textsc{IEEE Transactions On Circuits And Systems For Video Technology}, and the
\textsc{IEEE Transactions On Neural Networks And Learning Systems}.
\end{IEEEbiography}
\begin{IEEEbiography}[{\includegraphics[width=1in,height=1.25in,clip,keepaspectratio]{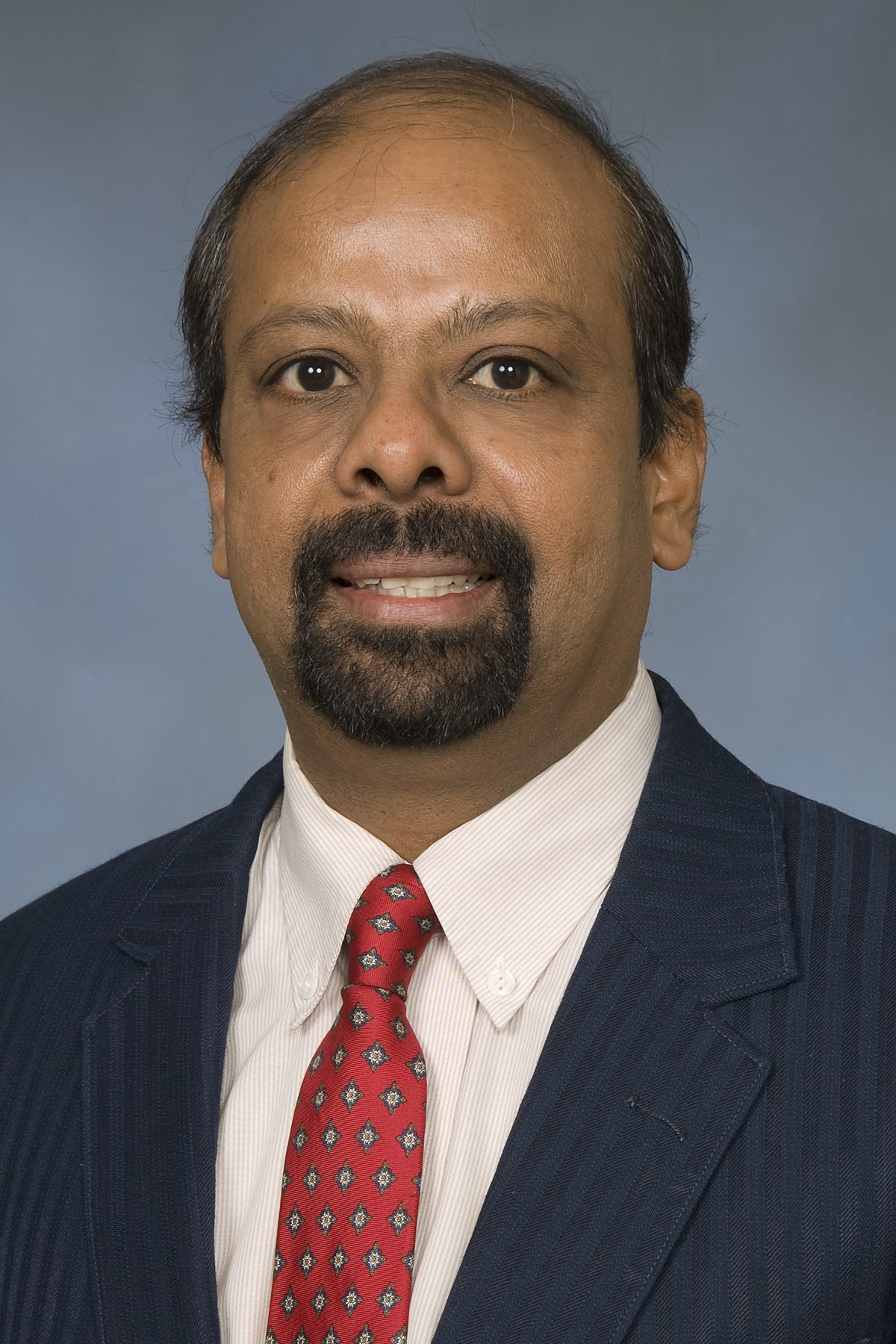}}]{Raghuveer M. Rao}(Fellow, IEEE)  received the M.E. degree in electrical communication engineering from
the Indian Institute of Science, Malleswaram, Bangalore, India, and the Ph.D. degree in electrical engineering from the University of Connecticut, Storrs,CT, USA, in 1981 and 1984, respectively.
He was a Member of Technical Staff at AMD
Inc., Santa Clara, CA, USA, from 1985 to 1987.
He joined the Rochester Institute of Technology,
Rochester, NY, USA, in December 1987, where,
at the time of leaving in 2008, he was a Professor of electrical engineering and imaging science. He is currently the Chief of the Intelligent Perception Branch, U.S. DEVCOM Army Research Laboratory (ARL), Adelphi, MD, USA, where he manages researchers and programs in computer vision and scene understanding. He has held visiting appointments at the Indian Institute of Science, the Air Force Research Laboratory, Dahlgren, VA, USA, the Naval Surface Warfare Center, Rome, NY, USA, and Princeton University, Princeton, NJ, USA. His research contributions cover multiple areas, such as signal processing with higher order statistics, wavelet transforms and scale-invariant systems, statistical self-similarity, and their applications to modeling in communication and image texture synthesis. His recent work is focused on machine learning methods for scene understanding on maneuvering
platforms. Dr. Rao is an elected fellow of the \textsc{International Society for Optical Engineers (SPIE)}. He is an ABET program evaluator for electrical engineering and has served on the Technical Committee for the \textsc{Ieee Signal Processing Society} and \textsc{SPIE}. He has also served as an Associate Editor for the \textsc{Ieee
Transactions On Signal Processing, the Ieee Transactions On Circuits And Systems}, and \textsc{The Journal of Electronic Imaging}.
\end{IEEEbiography}
\end{document}